%% file: main.tex
\documentclass[sigconf,screen]{acmart}
\usepackage{booktabs}
\usepackage{hyperref}
\usepackage{multirow}
\usepackage{xspace}
\usepackage{seqsplit} 
\usepackage{soul} 
\usepackage{enumitem} 
\usepackage[labelformat=simple]{subcaption} 
\usepackage{array} 
\usepackage[table]{xcolor} 
\usepackage{wrapfig} 
\usepackage{amsmath} 
\usepackage[linesnumbered,ruled,vlined]{algorithm2e}
\usepackage{siunitx}   

\SetKwProg{Fn}{Function}{}{:}{}

\usepackage[]{collab}
\collabAuthor{zb}{teal}{Zhuangbin}
\collabAuthor{zx}{purple}{Zhixiang}

\newcommand{\MethodName}{IcFuzz\xspace}
\newcommand{\SimName}{Isaac Sim\xspace}

\newcommand{\blue}[1]{#1}
\AtBeginDocument{%
  }

\setcopyright{cc}
\setcctype{by-nc-nd}
\acmDOI{10.1145/3832783.3837550}
\acmYear{2026}
\copyrightyear{2026}
\acmISBN{979-8-4007-2882-2/2026/10}
\acmConference[ASE '26]{Proceedings of the 41st IEEE/ACM International Conference on Automated Software Engineering}{October 12--16, 2026}{Munich, Germany}
\acmBooktitle{Proceedings of the 41st IEEE/ACM International Conference on Automated Software Engineering (ASE '26), October 12--16, 2026, Munich, Germany}
\acmSubmissionID{ase26main-p3261-p}
\received{2026-03-26}
\received[accepted]{2026-06-18}

\makeatletter
\AtEndPreamble{%
  \global\@ACM@balancefalse
  \RequirePackage{pbalance}
}
\makeatother
\begin{document}

\title{IcFuzz: Fuzzing Isaac Sim with Semantic Stage Guidance and Multi-level Mutation}

\settopmatter{authorsperrow=4} 

\author{Zhixiang Chen}
\authornotemark[2]
\orcid{0009-0002-6536-5978}
\affiliation{%
  \institution{Sun Yat-sen University}
  \city{Zhuhai}
  \country{China}
}
\email{chenzhx69@mail2.sysu.edu.cn}

\author{Zhuangbin Chen}
\authornote{Zhuangbin Chen is the corresponding author.}
\authornote{Zhixiang Chen, Zhuangbin Chen, Ruoxi Jia, Wei Li, and Zibin Zheng are also with the Zhuhai Key Laboratory of Trusted Large Language Models, Sun Yat-sen University, Zhuhai, China.}
\orcid{0000-0001-5158-6716}
\affiliation{%
  \institution{Sun Yat-sen University}
  \city{Zhuhai}
  \country{China}
}
\email{chenzhb36@mail.sysu.edu.cn}

\author{Ruoxi Jia}
\authornotemark[2]
\orcid{0009-0003-6294-2124}
\affiliation{%
  \institution{Sun Yat-sen University}
  \city{Zhuhai}
  \country{China}
}
\email{jiarx3@mail2.sysu.edu.cn}

\author{Zeqin Liao}
\orcid{0000-0003-0306-7465}
\affiliation{%
  \institution{Nanyang Technological University}
  \city{Singapore}
  \country{Singapore}
}
\email{zeqin.liao@ntu.edu.sg}

\author{Wei Li}
\authornotemark[2]
\orcid{0009-0002-1333-4408}
\affiliation{%
  \institution{Sun Yat-sen University}
  \city{Zhuhai}
  \country{China}
}
\email{liwei378@mail2.sysu.edu.cn}

\author{Jinyang Liu}
\orcid{0000-0003-0037-1912}
\affiliation{%
  \institution{Chinese University of Hong Kong}
  \city{Hong Kong}
  \country{China}
}
\email{jyliu@cse.cuhk.edu.hk}

\author{Zibin Zheng}
\authornotemark[2]
\orcid{0000-0002-7878-4330}
\affiliation{%
  \institution{Sun Yat-sen University}
  \city{Zhuhai}
  \country{China}
}
\email{zhzibin@mail.sysu.edu.cn}

\renewcommand{\shortauthors}{Zhixiang Chen, Zhuangbin Chen, Ruoxi Jia, Zeqin Liao, Wei Li, Jinyang Liu, and Zibin Zheng}

\begin{abstract}
Robotics simulators serve as a foundational infrastructure for embodied AI, facilitating safe and scalable robotic system development.
NVIDIA Isaac Sim has emerged as one of the most popular simulators, distinguished by its GPU-accelerated physics engine and photorealistic rendering, which enable high-fidelity modeling of complex environments.
However, its inherent complexity inevitably introduces software bugs that can compromise simulation reliability. 
Existing fuzzing approaches struggle to test Isaac Sim effectively due to challenges of context-aware object semantics, hierarchical simulation control, and a vast simulation state space.

In this paper, we propose \MethodName, the first fuzzing approach for Isaac Sim.
\MethodName first performs an LLM-based semantic stage segmentation, decomposing simulation programs into structured stages that capture context-aware object semantics. Guided by this information, \MethodName designs multi-level mutation operators to systematically exercise the simulator across hierarchical granularities.
To efficiently navigate the vast simulation state space, \MethodName employs a multi-armed bandit algorithm to adaptively schedule mutation operators.
Experimental results show that \MethodName outperforms the baselines in terms of both code coverage and bug detection.
Specifically, \MethodName achieves approximately 190\%--205\% of the code coverage of the baselines and detects an average of 3.7 unique crashes over three rounds of 12-hour tests, while no crashes are detected by the baselines.
Moreover, \MethodName has uncovered 11 \blue{bugs} over approximately four months, 9 of which have been confirmed or fixed by the developers.

\end{abstract}



\maketitle

\section{Introduction}
\label{sec:intro}
\input{section/01-introduction}

\section{Background and Motivation}
\label{sec:background}
\input{section/02-background}

\section{Methodology}
\label{sec:methodology}
\input{section/03-methodology}

\section{Evaluation}
\label{sec:evaluation}
\input{section/04-evaluation}

\section{Threats to Validity}
\label{sec:threats}
\input{section/05-threats}

\section{Related Work}
\label{sec:related}
\input{section/06-related_work.tex}

\section{Conclusion}
\label{sec:conclusion}
\input{section/07-conclusion}

\begin{acks}
This work was supported by the National Natural Science Foundation of China (No. 62402536).
\end{acks}

\section*{Data Availability}
We release our replication package at Zenodo~\cite{IcFuzz-GitHub-Repo}.

\bibliographystyle{ACM-Reference-Format}
\bibliography{refs}

\end{document}

%% file: section/01-introduction.tex
Robots are increasingly integrated into modern society, expanding their footprint across a variety of sectors~\cite{licardo2024intelligent, kyrarini2021survey, DBLP:journals/frobt/PietrantoniFFMMBDA24-Collaborative-Robots}. The global robotics market has reached \$50 billion and is projected to grow substantially in the coming decade~\cite{RoboticMarket}. 
Embodied AI has recently emerged as a key paradigm that endows robots with increasingly powerful capabilities by extending artificial intelligence from the digital world to physical systems~\cite{LinLiangSurvey-EmbodiedAI, DBLP:journals/tetci/DuanYTZT22-EmbodiedAI-Survey, DBLP:journals/arxiv/Bigazzi25-Autonomous-Embodied-Agents}.
Given the safety risks and prohibitive costs of direct experimentation in physical environments, robotics simulators have become the indispensable infrastructure for the advancement of embodied AI. They provide safe, efficient, and scalable platforms for training, testing, and validating robotic systems before real-world deployment~\cite{LinLiangSurvey-EmbodiedAI, DBLP:journals/tetci/DuanYTZT22-EmbodiedAI-Survey, DBLP:journals/arxiv/Bigazzi25-Autonomous-Embodied-Agents}.

Among existing robotics simulators, NVIDIA \SimName~\cite{IsaacSim-NvidiaPage} stands as one of the most influential and widely used platforms for embodied AI research and development, with adoption by over 100 companies (e.g., Siemens, Boston Dynamics, and Figure AI)~\cite{DBLP:conf/icse/ZhouSXS0LYS24-AICPSwithIsaacSim,IsaacSim-Adoption-NewsPage, IsaacSim-Adoption-NewsPage-Nvidia}. Distinguished by its GPU-accelerated physics engine and extensive library of digital assets, \SimName provides high-fidelity simulation and photorealistic rendering that accurately model real-world environments and complex physical interactions~\cite{LinLiangSurvey-EmbodiedAI,  IsaacSim-Tech-Blog-Photorealistic-Rendering, DBLP:journals/ral/MittalYYLRHYSGMMBSHG23-Orbit}. 
However, the inherent complexity of \SimName inevitably introduces software bugs that can compromise simulation reliability. 
For example, simulation crashes~\cite{Bug-Crash-at-training,Bug-Crash-at-random-moment} can force developers to expend substantial effort on troubleshooting, thereby undermining overall development efficiency. When such failures occur during model training or within operational digital twin systems, they may also lead to significant data loss and severe safety problems~\cite{DBLP:journals/arxiv/LiaoZRLGCLN25-EAIR-Bugs, DBLP:journals/aisoc/Viljanen24-Safety-by-Simulation, DBLP:conf/icra/CarlsonMN04-Mobile-Robot-Failures}. Therefore, testing \SimName is of paramount importance to ensure simulation reliability and to support the effective development of trustworthy embodied AI systems.

The systematic testing of \SimName necessitates effective automated testing techniques. To the best of our knowledge, there is currently no testing approach dedicated to \SimName. Fuzzing is a widely used method that exposes software defects by automatically generating random or malformed inputs~\cite{DBLP:journals/csur/MallisseryW24-Demystify-Fuzzing, DBLP:conf/sse/HuangZMC25-LLM-Fuzzing-Challenges, DBLP:journals/electronics/ZhangZZY23-Network-Protocol-Fuzzing}. It has been demonstrated to be a promising approach for uncovering previously unknown bugs in complex software systems~\cite{DBLP:journals/tse/ManesHHCESW21-Fuzzing-Survey, DBLP:journals/csur/MallisseryW24-Demystify-Fuzzing, DBLP:journals/cmc/YuLCLY24-Fuzzing-Survey}. 
However, the diverse functionality, sophisticated architecture, and complex input space of \SimName pose significant challenges to the application of fuzzing, hindering the generation of effective and diverse test cases. We summarize the main challenges as follows:

\begin{itemize}[noitemsep,leftmargin=5.5mm]
    \item \textbf{Context-aware Object Semantics.} The declaration of objects within \SimName is tightly coupled with their execution context, exhibiting complex contextual dependencies and strict logical ordering.
    For example, \textit{SimulationApp}~\cite{IsaacSim-NvidiaAPIDoc-SimulationApp} (which initializes the simulator runtime) must be instantiated prior to any other simulator-specific objects.
    Similarly, an \textit{Articulation}~\cite{IsaacSim-NvidiaAPIDoc-Articulation} can only be declared based on pre-existing \textit{prims}~\cite{OmniverseDoc-Prim} (basic scene elements identified by paths in the scene, such as a robot arm segment), since it is a high-level wrapper over one or more \textit{prims}.
    Inputs that violate these semantics (e.g., declaring an \textit{Articulation} before any \textit{prims} exist) are immediately rejected, thereby restricting the exploration of the simulator's internal logic. \looseness=-1
    \item \textbf{Hierarchical Simulation Control.} 
    Simulation control in \SimName is hierarchically structured across multiple granularities. 
    For example, adding a robot requires instantiating a specific object, such as a \textit{WheeledRobot}~\cite{IsaacSim-NvidiaAPIDoc-WheeledRobot} or a \textit{Manipulator}~\cite{IsaacSim-NvidiaAPIDoc-Manipulator}. Each type of object supports its distinct operations (e.g., setting wheel speeds for a \textit{WheeledRobot} or configuring joint positions for a \textit{Manipulator}). Each operation further requires its own parameters: wheel control requires separate velocities for each wheel, whereas joint control requires precise coordinate values. Control at different granularities exerts varied influences on the simulator. Consequently, systematic testing of Isaac Sim requires test generation across multiple levels of granularities. \looseness=-1
    
    \item \textbf{Vast Simulation State Space.} \SimName possesses a vast simulation state space, including diverse models (e.g., geometries, robots), complex physical interactions (e.g., collisions, lighting), and sophisticated AI workflows (e.g., reinforcement learning).
    The combinatorial explosion induced by these elements renders exhaustive exploration computationally infeasible.
    Random fuzzing typically struggles to reach deep or subtle states that may potentially contain bugs.  
    Therefore, effective strategies are required to guide fuzzing and ensure efficient exploration.
    
\end{itemize}

\looseness=-1
Given these challenges, existing fuzzing approaches demonstrate limited efficacy when applied to \SimName. 
General-purpose fuzzers (e.g., AFL++~\cite{DBLP:conf/woot/MaierEFH20-AFL++}, Atheris~\cite{Atheris}) generate test cases mainly at the level of flat byte streams. 
They either 1) mutate entire inputs (e.g., via byte insertion/deletion), which are often rejected due to violations of the simulator's format constraints and semantic dependencies, or 2) produce random parameters only for manually specified targets (e.g., functions), thereby enabling only single-granularity testing over limited functionality.
GzFuzz~\cite{DBLP:journals/pacmse/RenLLQXJ25-GzFuzz} is the state-of-the-art (SOTA) approach for fuzzing robotics simulators, specifically designed for Gazebo~\cite{DBLP:conf/iros/KoenigH04-Gazebo}. 
It fuzzes Gazebo via a command-line interface and leverages reinforcement learning to generate command sequences. These sequences primarily manipulate (e.g., adding, deleting, or moving) entire models (i.e., objects) and plugins (i.e., components providing custom functionalities) within simulation.
However, finer-grained simulation control is not captured (e.g., plugin-provided functionalities and their associated parameters) and therefore remains underexplored. For \SimName, which especially features a more complex architecture and richer functionality~\cite{IsaacSim-Tech-Blog-Photorealistic-Rendering}, such command sequences can explore only a small subset of its full capabilities~\cite{IsaacSim-NvidiaDoc-ROS2-SimulationControl}.
Consequently, both general-purpose fuzzers and GzFuzz cannot well handle all three challenges, which limits their ability to effectively explore the vast simulation state space of \SimName.

\looseness=-1
In this paper, we propose \MethodName, the first fuzzing approach for \SimName to the best of our knowledge.
\MethodName systematically tests the simulator by generating diverse and semantically valid simulation programs, which encompass the entire execution cycle of automated robotic simulations.
Our key insight is that these programs inherently follow a recurring semantic structure, which can be leveraged to constrain and guide mutation.
To address the challenge of context-aware object semantics, \MethodName explicitly identifies these structures as semantic stages, and then leverages an LLM to perform \textit{semantic stage segmentation}, decomposing each seed into code segments aligned with domain-specific execution stages.
Such a design is grounded in the official robot development lifecycle~\cite{IsaacSim-NvidiaDoc-Reference_Architecture} and reflects how developers typically construct simulation workflows in practice.
Building on this, \MethodName further utilizes a \textit{context-aware object selection} strategy, which restricts the set of candidate objects according to the corresponding semantic stage, preserving semantic integrity while maintaining object diversity.
To systematically exercise the hierarchical simulation control, \MethodName designs \textit{multi-level mutation} operators that combine program analysis and LLMs to mutate simulation objects, their supported operations, and associated arguments.
Furthermore, \MethodName employs a multi-armed bandit (MAB) algorithm to guide the scheduling of these mutation operators, enabling efficient exploration of the vast simulation state space in \SimName to maximize code coverage and bug detection.
\blue{In this work, \MethodName focuses on crash bugs as the test oracle, as they represent a severe and common class of failures in \SimName.}

\looseness=-1
We conduct a comprehensive evaluation of \MethodName and compare it against Atheris, a widely used general-purpose fuzzer for Python, and GzFuzz, the SOTA fuzzing approach for robotics simulators. 
Experimental results show that \MethodName achieves an average code coverage of 20,771 lines, which is approximately 205\% and 190\% of Atheris and GzFuzz, respectively. 
\MethodName detects an average of 7 crashes over three rounds of 12-hour tests (3.7 unique crashes after manual inspection), outperforming both baselines.
More importantly, \MethodName successfully reported 11 \blue{bugs}, 9 of which have been confirmed or fixed by developers. Ablation studies further validate the necessity and effectiveness of each key component within \MethodName. 

The main contributions of this work are summarized as follows:

\begin{itemize}[noitemsep,leftmargin=5.5mm]
    \item We propose \MethodName, the first fuzzing approach for \SimName. 
    \MethodName introduces semantic stage segmentation and context-aware object selection to handle the complex object semantics in simulation. It devises multi-level mutation operators to systematically exercise hierarchical simulation control across multiple granularities. Furthermore, \MethodName utilizes an MAB algorithm to schedule mutation operators for efficient exploration of the vast simulation state space in \SimName. \looseness=-1
    \item We present semantic stage segmentation as a novel semantic constraint mechanism to support the generation of semantically valid and diverse test cases.
    \item Experimental results show that \MethodName outperforms current SOTA baselines in code coverage and bug detection. \MethodName detected 11 \blue{bugs} within approximately four months, 9 of which have been confirmed or fixed. 
    \item We implement \MethodName and release the replication package at~\cite{IcFuzz-GitHub-Repo}.
\end{itemize}

%% file: section/02-background.tex
\begin{figure}
    \centering
    \includegraphics[width=0.87\linewidth]{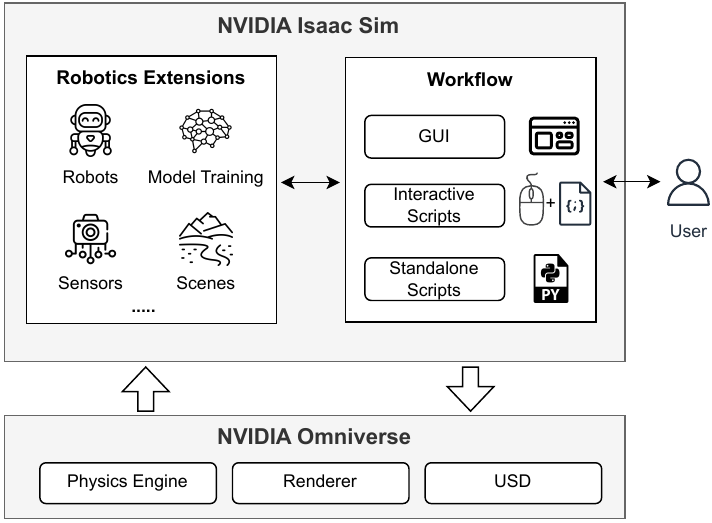}
    \caption{Overview of the \SimName architecture.}
    \label{fig:background_architecture}
    \vspace{-8pt}
\end{figure}
\textbf{Overview of \SimName.}
As shown in Fig.~\ref{fig:background_architecture}, \SimName is a robotics simulation platform built on the closed-source NVIDIA Omniverse~\cite{IsaacSim-NvidiaDoc-Omniverse}, which provides fundamental capabilities such as physics simulation, rendering, and scene representation based on Universal Scene Description (USD)~\cite{OpenUSD}. Built upon this foundation, \SimName primarily consists of a set of robotics-specific extensions, such as robots, sensors, scenes, and model training modules.
\blue{These extensions are open-source and encapsulate the core functionalities required for robotic simulation in \SimName.}
\blue{They} can be accessed separately through multiple workflows, namely a graphical user interface (GUI), interactive scripts, and standalone scripts~\cite{IsaacSim-NvidiaDoc-Workflow}.
The GUI provides a visual environment for developers to manipulate virtual scenes, such as assembling robots and attaching sensors.
Interactive scripts run asynchronously within the GUI, enabling on-demand script execution with real-time visual feedback.
Standalone scripts are complete Python programs that precisely control the simulation process (e.g., rendering steps) and constitute the primary modality for automated simulation tasks (e.g., model training and testing). Therefore, we adopt standalone scripts as the fuzzing input in this work.
Fig.~\ref{fig:semantic_stage} presents an example of a standalone \SimName script. \looseness=-1

\blue{The test subject of \MethodName is the entire \SimName simulator stack. During fuzzing, each input script exercises both the open-source extensions and the underlying closed-source Omniverse foundation. For coverage measurement, we collect coverage feedback only from the open-source \SimName extensions. As these extensions encapsulate the core functionalities required for robotic simulation, their coverage offers a meaningful signal for guiding the fuzzing process.} \looseness=-1

\textbf{Challenges in Fuzzing \SimName.}
Fuzzing \SimName is challenging due to the inherent characteristics of the simulator.
First, \SimName enforces \uline{Context-aware Object Semantics}, where object declarations are tightly coupled with the execution context and must follow strict semantic dependencies and logical ordering. 
Take Fig.~\ref{fig:semantic_stage} as an example, an \textit{Articulation} references prim paths of existing robots (e.g., ``\textit{Franka\_1}'' and ``\textit{Franka\_2}'') for unified control, while an \textit{IMUSensor} attaches to ``\textit{Franka\_1}'', illustrating prim-path dependencies.
During the interacting stage, joint positions and sensor data are accessed via the instantiated \textit{Articulation} and \textit{IMUSensor}, reflecting object dependencies. 
The chain-like logical ordering is enforced throughout the simulation lifecycle, e.g., the \textit{SimulationApp} must be initialized first, followed by the scene setup, before any objects can be meaningfully added or controlled. 
Inputs that violate these object semantics are rejected, resulting in shallow exploration. \looseness=-1

\begin{figure}
    \centering
    \includegraphics[width=0.87\linewidth]{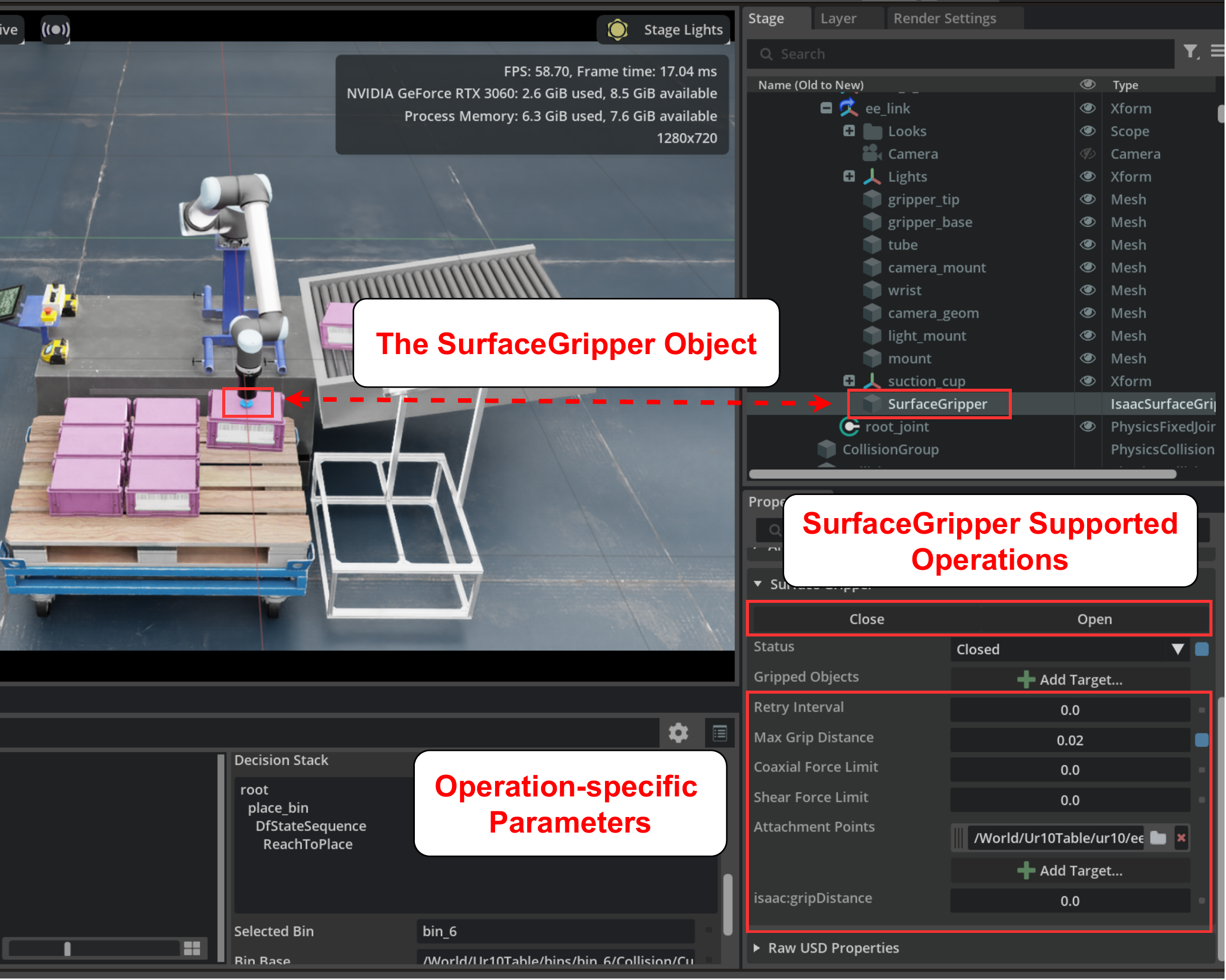}
    \caption{The \textit{SurfaceGripper} in \SimName.}
    \label{fig:background_snapshot}
    \vspace{-8pt}
\end{figure}

Simulation control in \SimName is hierarchically structured across multiple granularities. Fig.~\ref{fig:background_snapshot} shows a snapshot of a warehouse picking scenario, which involves diverse objects such as the \textit{SurfaceGripper} attached to a robotic arm, the pallet with boxes, etc. Each object supports its distinct operations. For instance, the \textit{SurfaceGripper} enables ``\textit{Close}'' for engaging grasp and ``\textit{Open}'' for release. 
Each operation requires specific parameters, such as the ``\textit{max grip distance}'' (the maximum allowable distance between the gripper and the target for a successful grasp).
\uline{Hierarchical Simulation Control} at these levels impacts the simulator differently, necessitating fully testing \SimName across different granularities.
Moreover, \uline{Vast Simulation State Space} in \SimName arises from the interplay of numerous objects, their supported operations, and operation-specific parameters. As a result, random fuzzing without appropriate guidance strategies tends to remain at shallow or repetitive states, making it difficult to reach deep or subtle states that may trigger bugs.

\begin{figure*}
    \centering
    \includegraphics[width=0.8\linewidth]{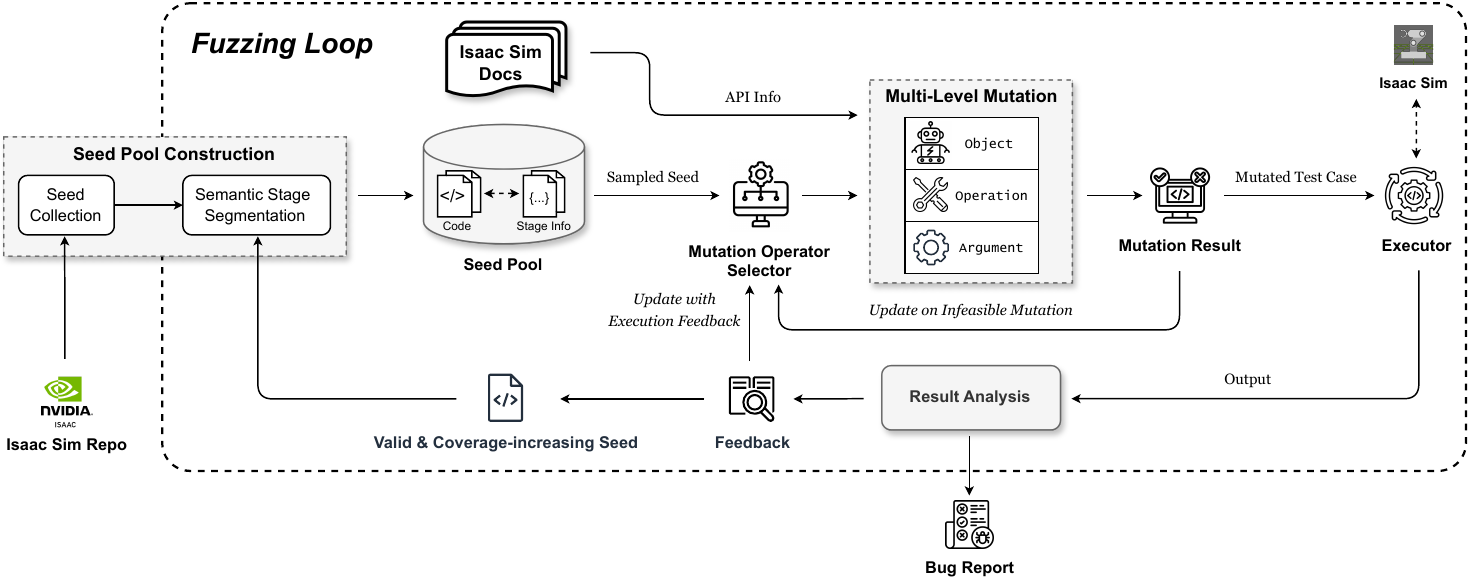}
    \caption{\blue{Overall Workflow of \MethodName.}}
    \label{fig:workflow}
\end{figure*}

Together, these challenges motivate \MethodName to generate semantically valid inputs, exercise simulation control at multiple granularities, and efficiently explore deep and diverse simulation states.


%% file: section/03-methodology.tex

Fig.~\ref{fig:workflow} illustrates the overall workflow of \MethodName. 
The key insight is that simulation programs inherently follow a structured lifecycle of semantic stages.
Based on this observation, \MethodName introduces \textit{semantic stage segmentation} (Sec.~\ref{sec:method-stage_segmentation}) to decompose each seed into code segments aligned with domain-specific execution stages.
The resulting stage annotations serve as explicit semantic constraints that guide context-aware object selection (Sec.~\ref{sec:method-object_selection}) during mutation, preserving semantic validity while maintaining diversity.

\looseness=-1
Given a sampled seed (Sec.~\ref{sec:method-seed_sampling}), \MethodName applies a multi-level mutation operator (Sec.~\ref{sec:method-mutation_operators}) that systematically exercises the hierarchical simulation control of Isaac Sim. 
The operators are adaptively scheduled based on an MAB algorithm~\cite{DBLP:journals/ml/AuerCF02-UCB} to efficiently explore the state space.
\MethodName assesses semantic feasibility during mutation generation, discarding infeasible results to reduce costly invalid executions while updating the selector. 
Finally, \MethodName executes the feasible mutated seed, analyzes the result, records any detected bugs, and updates the selector (Sec.~\ref{sec:method-exec_and_feedback}). 
Valid seeds that increase coverage undergo stage segmentation and re-enter the seed pool; this coverage-guided loop repeats until the time budget is reached.


\subsection{Seed Pool Construction}
We initialize the seed pool with valid Isaac Sim inputs, recording each seed alongside its segmented semantic stage information. 


\subsubsection{Seed Collection}
\label{sec:method-seed_collection}
\looseness=-1
\SimName can be driven by standalone Python scripts~\cite{IsaacSim-NvidiaDoc-Workflow}, which serve as individual test cases (see Sec.~\ref{sec:background}).
We collect all standalone scripts available in the official \SimName repository~\cite{IsaacSim-GitHubPage}. 
Some scripts require additional dependencies (e.g., robot model descriptions), which makes them difficult to run and mutate in isolation. 
As a result, we execute each script independently and retain only those that complete successfully without any errors. 
This process yields 116 seeds, which form a reliable set of initial test cases. 
These cases are broadly representative of real-world simulator usage and encompass a wide spectrum of functionalities.
They cover 20 distinct Isaac Sim packages (e.g., \textit{isaacsim.core}) across 8 major functional categories (e.g., robot manipulation), involving 9 robot platforms (e.g., Franka) and 6 sensor modalities (e.g., IMU), including tutorials and benchmarks maintained in the official repository.

\begin{figure}
    \centering
    \includegraphics[width=\linewidth]{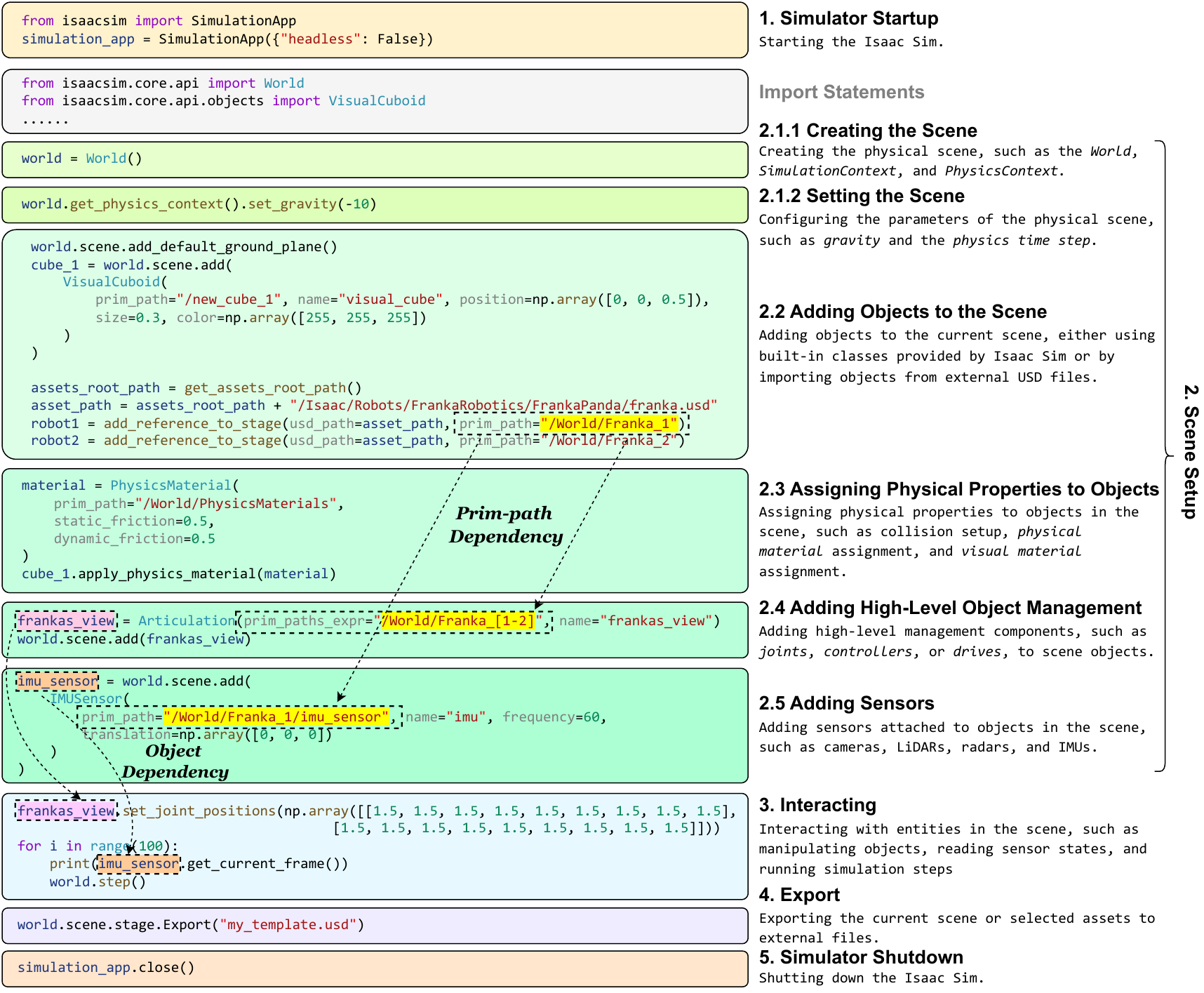}
    \caption{\blue{An example of semantic stage segmentation for a standalone Python script in \SimName.}}
    \label{fig:semantic_stage}
    \vspace{-8pt}
\end{figure}

\subsubsection{Semantic Stage Segmentation} 
\label{sec:method-stage_segmentation}
\looseness=-1
Object selection is crucial for fuzzing \SimName, as it directly influences the testing scope and the diversity of generated scenarios. 
However, valid object selection is challenging because object declarations must satisfy strict contextual dependencies and logical ordering constraints (see Sec.~\ref{sec:background}). 

Our key observation is that \SimName programs are not monolithic: they inherently follow a recurring \textit{semantic stage structure} that mirrors the robot development lifecycle. By explicitly identifying this structure, the open-ended problem of ensuring semantic validity can be reduced to a \textit{stage-constrained} problem.
Therefore, we propose segmenting each seed into semantic stages based on domain knowledge, which captures context-aware object semantics. 
Subsequent steps can leverage this information to constrain the set of permissible objects (see Sec.~\ref{sec:method-object_selection}), enabling semantic stage-guided mutations.
Code segmentation has demonstrated its effectiveness in other software engineering tasks such as code summarization~\cite{DBLP:conf/semco/SteinM23-CodeSegmeent-for-code-summary}, version identification~\cite{DBLP:conf/wcre/GerholdSZ24-CodeSegment-Version-Indentification}, and improving code readability~\cite{DBLP:conf/seke/DormuthGMS19-CodeSegment-readability,DBLP:journals/smr/WangPV14-CodeSegment-readability,DBLP:conf/profes/MayerMGSGP24-CodeSegment-readability}. \MethodName presents a novel use of this technique as a semantic constraint mechanism for fuzzing.



\looseness=-1
To derive a standardized and unified stage taxonomy, we systematically reviewed standalone scripts in the official Isaac Sim repository~\cite{IsaacSim-GitHubPage} and the reference architecture outlining the robot development lifecycle~\cite{IsaacSim-NvidiaDoc-Reference_Architecture}, to abstract common execution patterns.
Based on this analysis, we consolidate the most critical and representative operations into a set of semantic stages, as shown in Fig.~\ref{fig:semantic_stage}.
An \SimName test case starts by initializing \textit{SimulationApp} to launch the simulator before any simulator-specific import statements or operations (Stage 1).
Execution then proceeds to \textit{Scene Setup} (Stage 2), which includes creating the physical scene (Stage 2.1.1), configuring its parameters (Stage 2.1.2), and adding objects (Stage 2.2). 
The script then continues to \textit{Assigning Physical Properties} (Stage 2.3), \textit{Adding High-Level Object Management} (Stage 2.4), and \textit{Adding Sensors} (Stage 2.5) to scene objects.
Although the three stages (Stages 2.3-2.5) rely on pre-existing objects (e.g., an \textit{Articulation} groups two \textit{Franka robots} together), no strict ordering is imposed among them.
This design aligns with procedural flexibility in practice.
After completing \textit{Scene Setup} (Stage 2), the script further interacts with entities in the scene (Stage 3). 
Finally, the current stage is exported to a USD file (Stage 4) and the simulator is shut down (Stage 5).

\looseness=-1
As LLMs have demonstrated strong capabilities in code understanding~\cite{DBLP:conf/icse/NamMHVM24-code_understanding,DBLP:conf/icsm/RichardsW24-code_understanding,DBLP:journals/pacmse/KhojahM0N24-code_understanding}, we leverage them to automatically segment each seed into its semantic stages.
We design a chain-of-thought prompt~\cite{DBLP:conf/nips/Wei0SBIXCLZ22-CoT} that contains the complete seed, descriptions of each semantic stage, and an example output, instructing an LLM for accurate segmentation (full prompt available at~\cite{IcFuzz-GitHub-Repo}). Both the seed and its segmented stage information are finally stored in the seed pool.


\subsubsection{Seed Sampling}
\label{sec:method-seed_sampling}
\MethodName samples a seed at the beginning of each fuzzing iteration. To ensure sufficient exploration of the seed pool and avoid repetitive selection of certain seeds, we follow the seed sampling strategy in~\cite{DBLP:conf/sigsoft/WangYCLZ20-Lemon}. 
Each seed is assigned a weight inversely proportional to its historical sampling frequency, and these weights are normalized to determine the selection probability. Thus, frequently selected seeds have lower chances of being chosen, while less frequently sampled seeds are prioritized.
Newly added seeds during fuzzing can also be sampled for further mutations.

%
\begin{figure}
    \centering
    \includegraphics[width=0.85\linewidth]{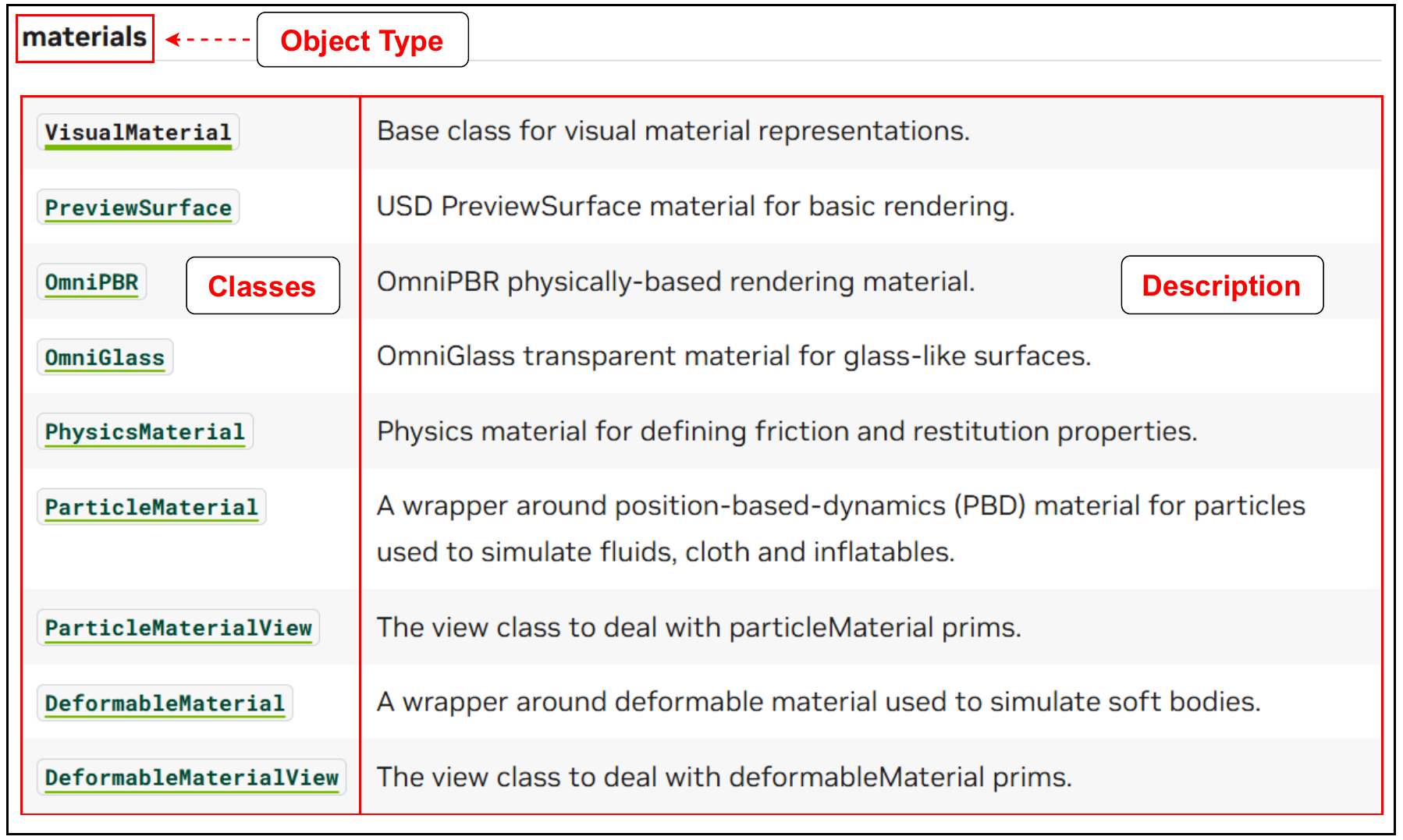}
    \caption{An excerpt from the \SimName documentation~\cite{IsaacSim-NvidiaAPIDoc-CoreAPI}.}
    \label{fig:api_object_type_material}
    \vspace{-10pt}
\end{figure}

\subsection{Multi-level Mutation}
\label{sec:method-multi_level_mutation}



\looseness=-1
We design multi-level mutation operators that act on seeds at the object, operation, and argument levels to systematically test Isaac Sim.


\subsubsection{Context-aware Object Selection}
\label{sec:method-object_selection}
Among the three levels of mutation, object-level mutation is the most critical. By altering the composition of objects in the scene, it largely shapes the testing scope and scenario diversity, and thus forms the foundation for the other two levels of mutation.
However, as discussed in Sec.~\ref{sec:background} and Sec.~\ref{sec:method-stage_segmentation}, object selection is non-trivial: it requires determining which objects can be validly added to a scene and where they can be introduced in the code while preserving semantic correctness.
To this end, we identify a stage-specific set of permissible objects for each semantic stage, thereby mitigating the complexity arising from context-aware object semantics. 


\looseness=-1
\SimName groups classes with similar functionalities into distinct categories, which we refer to as \textit{object types}. 
For example, Fig.~\ref{fig:api_object_type_material} illustrates various classes under the \textit{Materials} category, which can be instantiated as concrete objects and applied to prims to endow them with different physical properties.
We systematically reviewed all object types within the documentation to identify those eligible for addition to a scene or replacement of existing ones at each stage, where three authors independently annotated the mappings, clarified ambiguous cases by executing minimal scripts, and resolved inconsistencies through discussion. The results are presented in Table~\ref{tab:stage_object_mapping}.
For instance, during the \textit{Simulator Startup} stage, no additional objects are expected to be introduced. Therefore, both entries in the table are left empty. In contrast, at the \textit{Assigning Physical Properties} stage, any class in the \textit{Materials} category can be instantiated and then either added to the scene or used to replace an existing material object.
During object-level mutation, we select a target object by randomly sampling from the set of permitted objects in Table~\ref{tab:stage_object_mapping}, according to the applied mutation operator and the semantic stage. 


This design is motivated by the observation that seed scripts are already semantically valid and typically follow these semantic stages. Therefore, applying functionally similar mutations within the same stage (e.g., replacing one material with another) is likely to preserve semantic validity, as the preceding stages have already established the required context (e.g., objects are added to the scene before physical properties are assigned).



\begin{table}
\caption{Addable and replacement object types across different semantic stages in \SimName.}
\centering
\resizebox{\linewidth}{!}
{
    \begin{tabular}{l|l|l}
    \specialrule{0.35mm}{0em}{0em}
    \textbf{Semantic Stages}                    & \textbf{Addable Object Types}                                                                                       & \textbf{Replacement Object Types}                                                                                    \\ 
    \specialrule{0.15mm}{0em}{0em}
    \specialrule{0.15mm}{.1em}{0em}
    1. Simulator Startup              & —                                                                                                                   & —                                                                                                                    \\ 
    \hline
    2.1.1 Creating the Scene          & —                                                                                                                   & —                                                                                                                    \\ 
    \hline
    2.1.2 Setting the Scene           & —                                                                                                                   & —                                                                                                                    \\ 
    \hline
    2.2 Adding Objects to the Scene   & \begin{tabular}[c]{@{}l@{}}Objects, Robots, Cloners, Lights,\\Meshes, Shapes, DataLogger\end{tabular}               & \begin{tabular}[c]{@{}l@{}}Objects, Robots, Lights, Meshes,\\Shapes\end{tabular}                                     \\ 
    \hline
    2.3 Assigning Physical Properties & Materials                                                                                                           & Materials                                                                                                            \\ 
    \hline
2.4 High-Level Object Management  & \begin{tabular}[c]{@{}l@{}}Robots, Controllers, Grippers,\\Single Prims, Prims, Manipulators,\\Wrappers\end{tabular} & \begin{tabular}[c]{@{}l@{}}Robots, Controllers, Grippers,\\Single Prims, Prims, Manipulators,\\Wrappers\end{tabular}  \\ 
    \hline
    2.5 Adding Sensors                & Sensors                                                                                                             & Sensors                                                                                                              \\ 
    \hline
    3. Interacting                    & —                                                                                                                   & —                                                                                                                    \\ 
    \hline
    4. Export                         & Writers                                                                                                             & Writers                                                                                                              \\ 
    \hline
    5. Simulator Shutdown             & —                                                                                                                   & —                                                                                                                    \\
    \specialrule{0.35mm}{0em}{0em}
    \end{tabular}
}
\vspace{-8pt}
\label{tab:stage_object_mapping}
\end{table}

\subsubsection{Mutation Operators}
\label{sec:method-mutation_operators}

\looseness=-1
We begin by crawling the full documentation of all classes in Isaac Sim, including their methods and associated object-type information, and store them in a database for use during mutation.
The mutation operators integrate program analysis with LLMs to generate context-consistent code.
For operators that require selecting an existing object from a seed for mutation, we adopt a \textbf{program-analysis-based seed object selection} strategy. Specifically, we extract the abstract syntax tree (AST) of the seed and analyze import statements, constructor invocations, and variable assignments to obtain the objects available either in a specific semantic stage or across the entire seed. Each extracted object is queried against the \SimName documentation, and we randomly sample an object from those that can be successfully found. This ensures that the selected object is relevant to \SimName, rather than a trivial one. 
\blue{For object replacement and deletion, we further protect simulation-essential objects: \textit{SimulationApp}, \textit{World}, and \textit{AppFramework} are not selected as mutation targets.}
Note that this seed object selection identifies an existing object already present in the seed, whereas \textit{context-aware object selection} in Sec.~\ref{sec:method-object_selection} determines suitable new objects from the documentation to be introduced into the seed.


\textbf{Object-Level Mutation Operators}. 
Object-level mutation modifies the composition of objects in the scene. We define three operators to explore diverse objects and their compositions.

\begin{figure}
    \centering
    \includegraphics[width=0.92\linewidth]{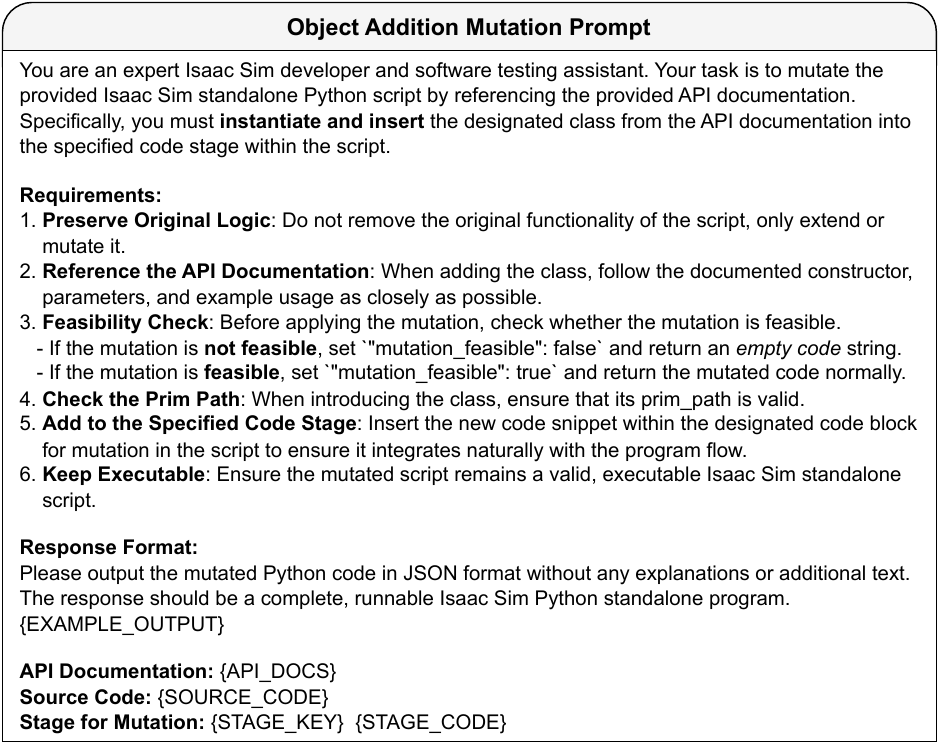}
    \caption{\blue{Simplified prompt for object addition mutation.}}
    \label{fig:object_addition_prompt}
    \vspace{-10pt}
\end{figure}

\begin{itemize}[noitemsep,leftmargin=5.5mm]

    \item \textbf{Addition} inserts a new object into a seed. 
	Based on the seed's semantic stage information, we randomly select a target stage from those both present in the seed and associated with addable object types (i.e., Stages 2.2–2.5 and Stage 4). 
	We then select a target object via \textit{context-aware object selection} (see Sec.~\ref{sec:method-object_selection}) and retrieve the constructor of its class from the documentation.
	Finally, we compose a prompt (see Fig.~\ref{fig:object_addition_prompt}) and feed it to an LLM to instantiate the target object and insert its code into the seed.
    
    \item \textbf{Replacement} replaces an existing object with another object. First, we select the target stage for mutation in the same manner as in \textit{addition}, except that we focus on replacement object types when referring to Table~\ref{tab:stage_object_mapping}. 
    We then select the original object from the target stage using the \textit{program-analysis-based seed object selection} strategy, and a replacement object via the \textit{context-aware object selection strategy} (see Sec.~\ref{sec:method-object_selection}). 
    Finally, we compose a prompt and feed it to an LLM. Different from \textit{addition}, the LLM is also responsible for replacing dependencies associated with the original object (e.g., method invocations), or removing them when replacement is infeasible.
    
    \item  \textbf{Deletion} removes an object from a seed. It follows the \textit{program-analysis-based seed object selection} to identify the target object, and instructs an LLM to remove it along with its dependents. 
    
\end{itemize}

\textbf{Operation-Level Mutation Operators}. 
\looseness=-1
Operation-level mutation mutates the supported operations of existing objects, which are concretely manifested as method invocations in seed scripts.  
We define three operators to explore diverse operations scattered across different objects, as well as their invocation sequences.
All three operators begin by selecting a target object via the \textit{program-analysis-based seed object selection} strategy. 
\blue{For replacement and deletion, the \textit{close} methods of \textit{SimulationApp} and \textit{AppFramework} are excluded from target selection to ensure proper simulation termination.}

\begin{itemize}[noitemsep, leftmargin=5.5mm]

    \item \textbf{Addition} randomly samples a method from the documentation of the selected object's class, then constructs a prompt incorporating the seed and the documentation snippet of the sampled method, and feeds it to an LLM to generate the insertion.
    
    \item \textbf{Replacement} follows the same method-sampling procedure as \textit{addition}, then selects an existing invocation of the target object and instructs an LLM to replace it with the sampled method. \looseness=-1
    
    \item \textbf{Deletion} randomly selects an existing invocation of the target object and instructs an LLM to remove it from the seed.

\end{itemize}

\looseness=-1
\textbf{Argument-Level Mutation Operators.} 
Argument-level mutation mutates the arguments of object operations, representing the finest mutation granularity. 
Given a seed, we select a target object via the \textit{program-analysis-based seed object selection} strategy, then randomly choose one of its methods that accepts at least one argument. We collect each parameter's type, default value, and nullability from the documentation. 
Each argument is mutated with 50\% probability using a randomly chosen operator.
All argument-level mutations are performed directly on the AST, without using LLMs. 

\begin{itemize}[noitemsep,leftmargin=5.5mm]
    \item \textbf{Deletion} removes a selected argument from the method call. This operator is applicable only if the argument has a documented default value. 
    
    \item \textbf{Nullification} sets the value of a selected argument to \texttt{None}, which can be applied only if the documentation indicates that the argument allows \texttt{None} as a valid value. 
    
    \item \textbf{Replacement} modifies the value of a selected argument. This operator is applicable only to arguments of numeric types (e.g., int, float), boolean type (i.e., bool), or lists/arrays containing numeric/boolean elements. For numeric arguments/elements, a new value is randomly generated to replace the original value. For boolean arguments/elements, the value is negated. 
\end{itemize}






\textbf{Mutation Feasibility Analysis.} 
Executing a test case in \SimName is time-consuming. For example, fully loading a warehouse sample scene can take more than 30 seconds~\cite{IsaacSim-NvidiaDoc-Benchmark_Time_Comsuming}. Therefore, we analyze the feasibility of each candidate mutation and discard infeasible ones when applying mutation operators, thereby reducing the time wasted on executing invalid test cases.

Specifically, we perform two types of feasibility analyses: 
\textbf{1) Rule-based analysis.} This analysis checks whether a seed satisfies the prerequisites of the selected mutation operator. For example, object addition/replacement operators require the seed to contain a semantic stage in which the set of addable/replaceable objects is non-empty (see Table~\ref{tab:stage_object_mapping}), whereas argument-level mutation operators require the seed to include \SimName-specific method invocations with arguments of mutable types. The mutation is deemed infeasible when the prerequisites are not met. 
\textbf{2) LLM-based analysis.} For mutation operators that rely on LLM-based generation (i.e., object- and operation-level mutation operators), we instruct the LLM to determine whether a semantically valid script (i.e., one that can be executed successfully without errors) can be produced under the given mutation instruction. 
\blue{Leveraging their strong code understanding capabilities, LLMs serve as a practical choice to flexibly reason about the simulator's dynamic semantics and contextual compatibility, avoiding the laboriously crafted rules required by traditional program analysis.}
An example prompt is shown in Fig.~\ref{fig:object_addition_prompt}. 
For instance, the LLM may reject a mutation when the object to be added is incompatible with the existing context of the seed. 
If either type of feasibility analysis determines that a mutation is infeasible, we discard it without generating the mutated seed.

\begin{algorithm}[t]
\footnotesize
\caption{UCB-based Mutation Operator Selection}
\label{alg:ucb-mutation-selection}

\KwIn{Mutation operator set $\mathcal{O} = \{o_1, \dots, o_K\}$}

\Fn{Initialize()}{ \label{line:ucb_initial_begin}
    \ForEach{operator $o_i \in \mathcal{O}$}{
        $n_i \gets 0$ \tcp*[r]{selection count}
        $R_i^{\text{sum}} \gets 0$ \tcp*[r]{cumulative reward}
        $\text{UCB}_i \gets +\infty$ \tcp*[r]{force initial exploration}
    }
    $N \gets 0$ \tcp*[r]{total selections} \label{line:ucb_initial_end}
}

\Fn{SelectOperator()}{ \label{line:ucb_select_op_begin}
    $i^\star \gets \arg\max_i \text{UCB}_i$\;
    \Return $o_{i^\star}$\; \label{line:ucb_select_op_end}
}

\Fn{Update($o_i, \text{NewCoverageGain}, \text{CrashFound}$)}{ \label{line:ucb_update_begin}
    $R \gets w_{\text{cov}} \times \text{NewCoverageGain} + w_{\text{crash}} \times \text{CrashFound}$\; \label{line:ucb_reward_function}
    $N \gets N + 1; \quad n_i \gets n_i + 1$; \quad
    $R_i^{\text{sum}} \gets R_i^{\text{sum}} + R$; \quad $\bar{R}_i \gets R_i^{\text{sum}} / n_i$\;
    $\text{UCB}_i \gets \bar{R}_i + \sqrt{\frac{2 \ln N}{n_i}}$\; \label{line:ucb_update_end}
}
\end{algorithm}

\textbf{Mutation Operator Selection.} 
Selecting mutation operators is a sequential decision-making problem under uncertainty, where operator effectiveness is only revealed after costly executions. 
With the goal of maximizing code coverage and bug detection under a limited budget, the selector must balance exploring less-used operators and exploiting those with demonstrated effectiveness. This trade-off naturally motivates modeling the selection as an MAB problem. \looseness=-1

Specifically, we formulate the mutation operator selection as a $K$-armed bandit problem. The set of mutation operators $\mathcal{O} = \{o_1, \dots, o_K\}$ corresponds to the $K$ distinct arms. The fuzzing process evolves over a sequence of discrete time steps $t = 1, 2, \dots, N$. At each step $t$, the selector chooses an operator $o_i$ to apply to a seed, an action equivalent to pulling the $i$-th arm of the bandit. The execution outcome serves as the stochastic reward, with the objective of maximizing the cumulative reward over the total selection horizon.

We solve this problem using the Upper Confidence Bound (UCB) algorithm~\cite{DBLP:journals/ml/AuerCF02-UCB}, as detailed in Algorithm~\ref{alg:ucb-mutation-selection}. In each iteration, we select the operator $o_{i^\star}$ with the highest UCB score (Lines~\ref{line:ucb_select_op_begin}--\ref{line:ucb_select_op_end}), which represents the largest upper confidence bound on the expected reward of the operator.
The reward function $R$ (Line~\ref{line:ucb_reward_function}) is computed as a weighted sum of coverage expansion and crash discovery:
\begin{equation}
    R = w_{\text{cov}} \times \text{NewCoverageGain} \;+\; w_{\text{crash}} \times \text{CrashFound},
\end{equation}
where NewCoverageGain and CrashFound are binary indicators: NewCoverageGain is set to 1 when the mutated seed is valid (i.e., executes successfully without errors) and increases cumulative code coverage, and CrashFound is set to 1 when a crash is triggered.
The UCB score (Line~\ref{line:ucb_update_end}) of the selected operator is updated as follows:  
\begin{equation}
    \text{UCB}_i = \bar{R}_i + \sqrt{\frac{2 \ln N}{n_i}},
\end{equation}
where $\bar{R}_i$ represents the empirical mean reward of operator $o_i$ (favoring exploitation), while the second term provides an exploration bonus that is positively correlated with the total number of selections 
$N$ and negatively correlated with the operator’s historical selection count $n_i$, ensuring that less-frequently selected operators still have a chance to be selected.

\subsection{Oracle and Feedback}
\label{sec:method-exec_and_feedback}
\MethodName executes mutated test cases in \SimName and analyzes their execution results to identify potential bugs.
Regarding the test oracle, we focus on crash bugs occurring during program execution, following prior work~\cite{DBLP:journals/pacmse/RenLLQXJ25-GzFuzz}. Such bugs cause abrupt termination of the simulation, potentially leading to loss of training data or safety risks, and therefore represent a severe class of failures. 
In addition, crash bugs constitute a common bug pattern in \SimName and are relatively easy to reproduce and localize, making them a practical entry point for improving the reliability of \SimName.
\blue{Non-crashing bugs, such as inaccurate sensor readings or violations of physical constraints, are not covered by the current oracle. Extending \MethodName to detect such bugs would require defining domain-specific semantic oracles (e.g., physics invariant checkers), instrumenting the simulator to collect richer runtime states, and incorporating semantic deviation signals into the feedback mechanism. We leave such extensions to future work.}

\MethodName leverages execution feedback to guide its fuzzing process. For each test case, \MethodName determines whether it executes successfully and measures the code coverage. A test case that completes without errors is considered valid. 
If a valid test case further increases coverage, it is subjected to semantic stage segmentation and added to the seed pool. 
Meanwhile, execution feedback, including validity, coverage gain, and crash occurrence, is fed back to the mutation operator selector.  
It updates the UCB score of each mutation operator accordingly to support adaptive scheduling.



%% file: section/04-evaluation.tex
In this section, we evaluate \MethodName to answer the following research questions (RQs):

\begin{itemize}[noitemsep,leftmargin=5.5mm]
    \item \textbf{RQ1 (Bug Detection)}: How does \MethodName perform in detecting \blue{bugs} for \SimName?
    \item \blue{\textbf{RQ2 (Effectiveness)}: How effective is \MethodName in its overall fuzzing performance and LLM-dependent intermediate steps?}
    \item \textbf{RQ3 (Ablation Study)}: What are the contributions of the key components in \MethodName?
\end{itemize}


\subsection{Experimental Setup}
\noindent\textbf{Baselines.}
\looseness=-1
We select GzFuzz~\cite{DBLP:journals/pacmse/RenLLQXJ25-GzFuzz} and Atheris~\cite{Atheris} as baselines.
GzFuzz is the SOTA approach for fuzzing robotics simulators and is designed for Gazebo. 
To apply it to \SimName, we make minimal modifications to its source code, primarily adapting its simulator interaction layer.
Atheris is a widely used general-purpose fuzzer for Python. Its fuzz testing is implemented by mutating input USD files, following the general-purpose fuzzer implementation in~\cite{DBLP:journals/pacmse/RenLLQXJ25-GzFuzz}.
Both fuzzers are run with their default settings.
\blue{Although grammar-based fuzzers and search-based test generation tools can produce structured test inputs, we do not include them as baselines.
Grammar-based fuzzing~\cite{DBLP:conf/icse/WangCWL19-Superion,DBLP:conf/ndss/AschermannFHJST19-NAUTILUS} generates or mutates inputs according to predefined rules, and adapting it to \SimName would require substantial expert effort to define input grammars; moreover, such grammars mainly capture syntactic structure, are hard to extend to simulator-specific semantics, and also limit test diversity.
Representative search-based tools (e.g., EvoSuite~\cite{DBLP:conf/sigsoft/FraserA11-EvoSuite} and Pynguin~\cite{DBLP:conf/icse/LukasczykF22-Pynguin}) mainly generate unit-level tests for classes, functions, or modules, whereas testing the entire \SimName requires interactions across multiple modules, and individual modules in \SimName are difficult to load and execute independently due to complex preconditions.
}

\noindent\textbf{Implementation of \MethodName.}
\MethodName is implemented in approximately 4,500 lines of Python code~\cite{IcFuzz-GitHub-Repo}.
We use SQLite~\cite{SQLite} to manage the seed pool and store the \SimName documentation.
\blue{By default, we use GPT-5-mini~\cite{OpenAI-GPT-5-mini} for semantic stage segmentation and mutation, and set the UCB reward weights to $w_{\text{cov}}=1$ and $w_{\text{crash}}=5$.}

\noindent\textbf{Environment.}
Experiments are conducted on a Windows 10 desktop with an Intel Core i5-13400F CPU at \SI{2.50}{GHz}, \SI{32}{GB} of RAM, and an RTX 3060 GPU.
RQ1 is evaluated on \SimName 5.0.0 and 5.1.0, while RQ2 and RQ3 use only version 5.1.0, the latest stable release. \looseness=-1


\begin{figure}
    \centering
    \includegraphics[width=\linewidth]{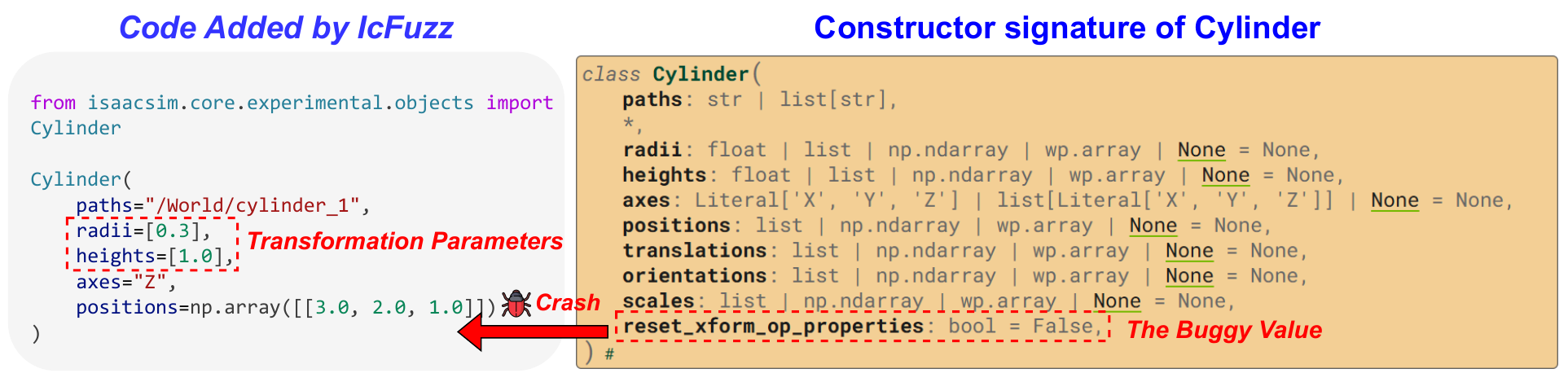} 
    \caption{\blue{A bug triggered by adding a \textit{Cylinder}.}}
    \label{fig:rq1_object_bug_case}
    \vspace{-11pt}
\end{figure}

\noindent\textbf{Metrics.} We detail the metrics used in our study as follows.

\begin{itemize}[noitemsep,leftmargin=5.5mm]
    \item 
    \textbf{Number of total/unique crashes.} We report total and unique crash counts. Crashes in \SimName exhibit explicit symptoms in the execution logs, such as messages indicating ``crash detected'' or the generation of dump files. \blue{For each detected crash, we collect its triggering script, exception type and message, and full stack trace. Three authors independently reproduce each crash in a clean \SimName environment and label two crashes as duplicates if they share the same exception type, top stack frames up to the triggering function, and triggering condition; otherwise, they are treated as distinct. We then compare the resulting labels and resolve any inconsistencies through joint discussion until consensus is reached.} Each unique crash is reported to the developers for further confirmation. \looseness=-1

    \item \textbf{Number of valid/invalid scripts.} A generated script is considered valid if it executes without errors (e.g., uncaught exceptions) as indicated by its logs; otherwise, it is invalid. We also count valid scripts that further increase cumulative code coverage (i.e., the number of valid and coverage-increasing scripts). \looseness=-1
    


    \item \textbf{Number of covered unique classes/replacement pairs.}
    This metric measures the number of unique classes (for object addition) or unique replacement pairs (for object replacement) that result in valid scripts after mutation. It reflects the diversity of objects effectively explored through object-level mutations.



    \item \textbf{Code coverage.} 
    Code coverage has been widely used in prior software testing studies~\cite{DBLP:conf/issta/Suo00JZW24-MLIR,DBLP:conf/sp/ParkKY25-RGFuzz,DBLP:journals/pacmse/RenLLQXJ25-GzFuzz}. We follow GzFuzz~\cite{DBLP:journals/pacmse/RenLLQXJ25-GzFuzz} and adopt line coverage as our coverage criterion, which is measured using \textit{coverage.py}~\cite{CoveragePy}. \blue{We use Welch's t-test with $p<0.05$ to determine statistical significance, following prior work~\cite{DBLP:conf/kbse/ZhaoWHX25-HFuzzer-t-test}.}
\end{itemize}




\begin{figure}
    \centering
    \includegraphics[width=0.75\linewidth]{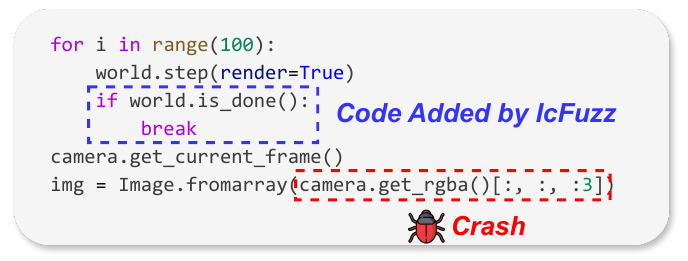}
    \caption{A bug triggered by an unexpected value returned by the camera.}
    \label{fig:rq1_method_bug}
    \vspace{-5pt}
\end{figure}

\begin{figure}
    \centering
    \includegraphics[width=0.8\linewidth]{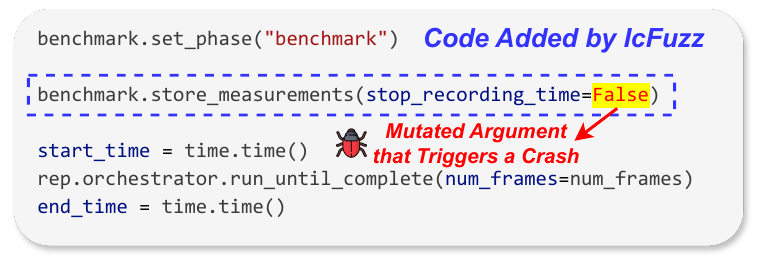}
    \caption{A bug triggered by an incompletely implemented argument.}
    \label{fig:rq1_arg_bug}
    \vspace{-9pt}
\end{figure}

\subsection{RQ1: Bug Detection}
To evaluate the effectiveness of \MethodName in detecting \blue{bugs}, we conducted a longitudinal fuzzing
campaign over approximately four months, following~\cite{DBLP:journals/pacmse/RenLLQXJ25-GzFuzz}. We tested the latest stable version of \SimName available at the time. Specifically, we initially evaluated \MethodName on version 5.0.0 and subsequently switched to 5.1.0 after its official release. 
During the study, 11 bugs are reported in total, among which 7 are confirmed by the developers (including 5 that are already fixed as of this writing), 2 are still awaiting further responses, and the remaining 2 are considered stale, as they had already been fixed in the latest developer branch at the time of reporting. Next, we present representative bug cases detected by \MethodName through object-, operation-, and argument-level mutations, respectively.

Fig.~\ref{fig:rq1_object_bug_case} illustrates \href{https://github.com/isaac-sim/IsaacSim/issues/280}{Bug \#280} as an example. The gray box shows the mutation code generated by \MethodName for adding a \textit{Cylinder} object to the scene, while the right side presents the constructor signature of the corresponding class~\cite{IsaacSim-NvidiaAPIDoc-Experimental_Object}. This bug is caused by an incorrect default value of the parameter \textit{``reset\_xform\_op\_properties''}, which is mistakenly set to False. As a result, when a \textit{Cylinder} object is added to the scene with transformation parameters (e.g., \textit{``radii''} and \textit{``heights''}), \SimName crashes immediately. 
Notably, this bug is not limited to the \textit{Cylinder} class; instead, it has a broad impact, affecting a total of 15 classes under the \textit{Shapes}, \textit{Lights}, and \textit{Meshes} categories.
These experimental classes, although expected to become the main classes in future releases~\cite{IsaacSim-NvidiaAPIDoc-CoreExperimentalAPI}, are only recently introduced and currently lack sufficient usage examples, and may also lack thorough testing by the project team.
The developers have confirmed this bug and created an internal ticket to address it.


Fig.~\ref{fig:rq1_method_bug} shows \href{https://github.com/isaac-sim/IsaacSim/issues/243}{Bug \#243}. 
\MethodName detected this bug by adding a call to the ``\textit{is\_done}'' method of the \textit{World} object and then prematurely terminating the simulation loop according to its return value.
This prevents the simulation from completing the necessary warmup steps before valid camera data is acquired.
As a result, ``\textit{get\_rgba}'' returns an unexpected one-dimensional array instead of the expected three-dimensional one, leading to out-of-bounds memory accesses and a crash.
This behavior contradicts the original documentation, which claims that ``\textit{get\_rgba}'' always returns an array with the correct dimensions. According to our suggestions, the developers have updated the documentation to alert users to potential output shape issues before the renderer has fully warmed up~\cite{IsaacSim-NvidiaAPIDoc-Camera}.


Fig.~\ref{fig:rq1_arg_bug} provides an overview of \href{https://github.com/isaac-sim/IsaacSim/issues/338}{Bug \#338}. 
\MethodName detected this bug through two rounds of mutation. First, it added the method call ``\textit{store\_measurements}'' into the seed. Subsequently, it mutated the value of the argument ``\textit{stop\_recording\_time}'' from \textit{True} to \textit{False} via argument replacement, which ultimately triggered the bug. 
After a month-long investigation, developers confirmed that the root cause was that the functionality for ``\textit{stop\_recording\_time=False}'' had not been fully implemented. Notably, directly calling ``\textit{store\allowbreak\_measurements}'' with the default value of ``\textit{stop\_recording\_time}'' as \textit{True} works correctly; the bug is only triggered when the argument is negated through mutation.

These case studies demonstrate that the triggering conditions of bugs in \SimName can be intricate and may involve non-trivial interactions across object-, operation-, and argument-level mutations. 
This observation further justifies the multi-level mutation of \MethodName, which systematically exercises the hierarchical simulation control of \SimName and effectively detects \blue{bugs}.


\subsection{\blue{RQ2: Effectiveness}}


\begin{figure}
    \centering
    \includegraphics[width=0.9\linewidth]{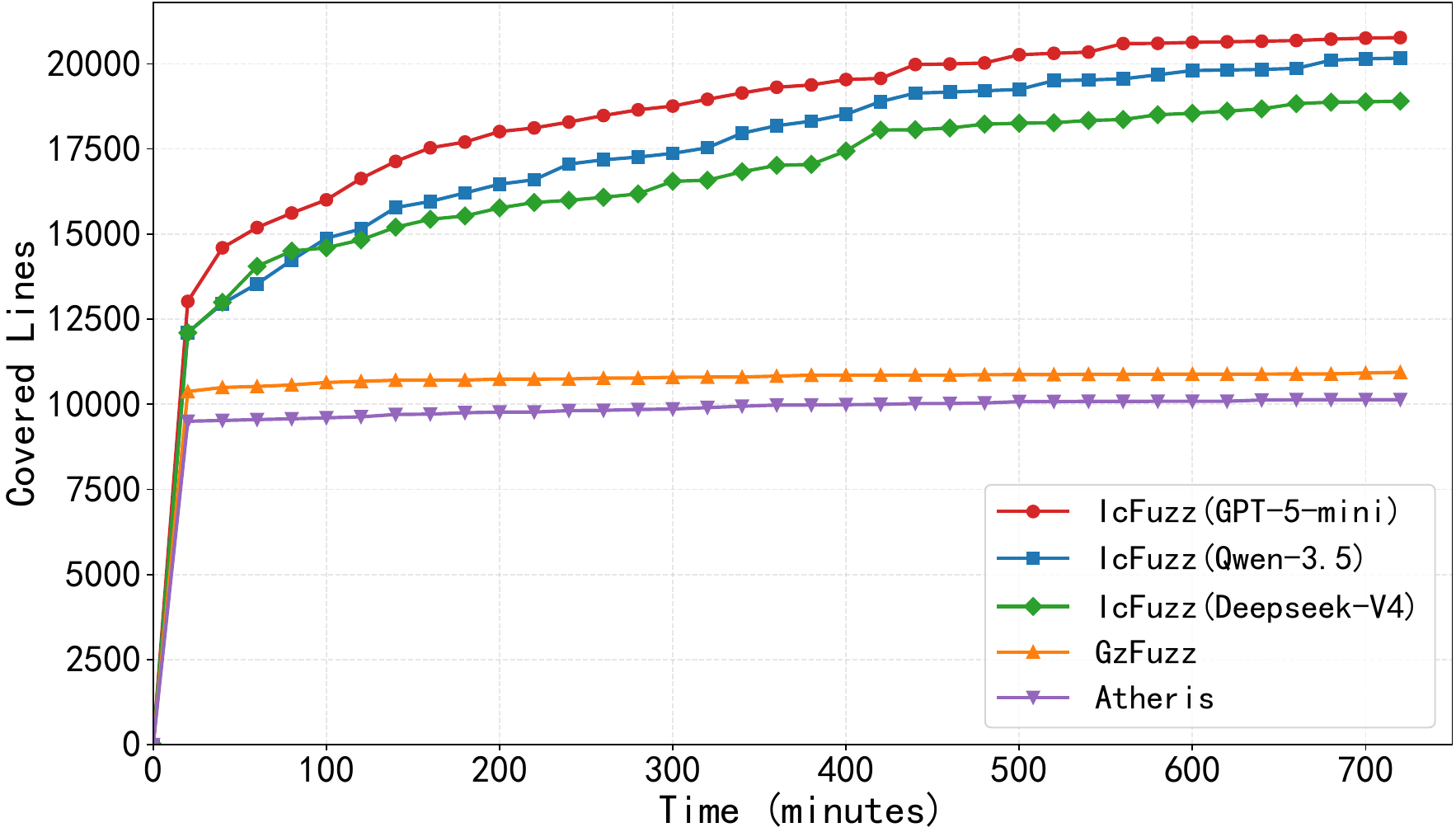} 
    \caption{\blue{Average code coverage over time.}}
    \label{fig:rq2_coverage}
    \vspace{-7pt}
\end{figure}

\begin{table}
\caption{\blue{Average crashes and coverage of \MethodName and baselines.}}
\centering
\resizebox{\linewidth}{!}
{
    \begin{tabular}{c|cccc} 
        \specialrule{0.35mm}{0em}{0em}
        \textbf{Method} & \textbf{\# Unique Crashes $\uparrow$} & \textbf{\# Total Crashes $\uparrow$} & \textbf{Coverage $\uparrow$} & \textbf{$p$-value}  \\ 
        \specialrule{0.15mm}{0em}{0em}
        \specialrule{0.15mm}{.1em}{0em}
        Atheris             & 0                       & 0                      & 10136.0            & 4.45E-5 \\ 
        \hline
        GzFuzz              & 0                       & 0                      & 10940.7            & 1.30E-6 \\ 
        \hline
        IcFuzz (DeepSeek-V4) & \textbf{4.3}                    & 5.3                    & 18903.3            & -- \\
        \hline
        IcFuzz (Qwen-3.5)    & 2.3                    & 2.7                    & 20167.7            & -- \\
        \hline
        IcFuzz (GPT-5-mini)              & 3.7            & \textbf{7.0}           & \textbf{20771.0}   & -- \\
            \specialrule{0.35mm}{0em}{0em}
    \end{tabular}
}
\label{tab:rq2_crashes_and_coverage}
\vspace{-11pt}
\end{table}

\subsubsection{\blue{Overall Effectiveness}}
\label{sec:evaluation-rq2_overall_effectiveness}
\looseness=-1
\blue{This section compares the performance of \MethodName with prior work and evaluates its overall effectiveness under different LLMs and UCB reward settings.}
\blue{For each \MethodName configuration and baseline, we conduct three independent 12-hour runs and report the average total/unique crashes and code coverage.} 

\textbf{\MethodName outperforms the baselines in both crash detection and code coverage.} 
\blue{The results achieved by \MethodName with the default LLM (GPT-5-mini) and by the baselines are shown in Table~\ref{tab:rq2_crashes_and_coverage}.} Atheris and GzFuzz fail to detect any crashes across all runs, whereas \MethodName detects an average of seven total crashes, including 3.7 unique crashes.
\MethodName also achieves the highest code coverage, covering an average of 20,771 lines of code, which is approximately 205\% of Atheris (10,136 lines) and 190\% of GzFuzz (10,940.7 lines); \blue{these differences are statistically significant.}
Fig.~\ref{fig:rq2_coverage} shows the coverage trends over time.
The coverage of GzFuzz and Atheris increases rapidly at the beginning but grows slowly after approximately 100 minutes. In contrast, \MethodName exhibits steady coverage growth from 0--400 minutes and does not plateau until around 600 minutes. 


\blue{To evaluate the overall effectiveness of \MethodName under different configurations, we replace the default LLM (GPT-5-mini) with DeepSeek-V4~\cite{DeepSeek-V4} and Qwen3.5-397B-A17B~\cite{Qwen3.5-397B-A17B}, and vary $w_{\text{crash}}$ among 1, 5 (default), and 10. The LLM variant results are included in Table~\ref{tab:rq2_crashes_and_coverage} and Fig.~\ref{fig:rq2_coverage}, and the reward setting results are shown in Table~\ref{tab:rq2_ucb_reward_settings}.
\textbf{The performance of \MethodName varies slightly across different LLMs and UCB reward settings, while all configurations substantially outperform the baselines.}
Across the three LLMs, code coverage ranges from 18,903.3 to 20,771.0 lines with 2.3--4.3 unique crashes detected on average. 
The lowest-coverage LLM configuration still achieves approximately 173\% of the code coverage of GzFuzz and 187\% of Atheris, and both baselines fail to trigger any crash.
}
\blue{As $w_{\text{crash}}$ decreases from 10 to 1, code coverage increases from 19,879.0 to 20,968.7 lines. With a lower $w_{\text{crash}}$, the UCB algorithm favors mutation operators that yield coverage gains, thus enabling broader code exploration. Meanwhile, increasing $w_{\text{crash}}$ does not proportionally improve crash detection, as crash occurrences are inherently sparse and stochastic, making them difficult to reliably increase by merely biasing operator selection. We set $w_{\text{crash}}=5$ as the default to balance coverage growth and crash detection.} \looseness=-1

\begin{table}
\caption{\blue{Average crashes and coverage of \MethodName under different UCB reward settings.}}
\centering
\resizebox{\linewidth}{!}
{
    \begin{tabular}{cc|ccc} 
        \specialrule{0.35mm}{0em}{0em}
        \textbf{$w_{\mathrm{cov}}$} & \textbf{$w_{\mathrm{crash}}$} & \textbf{\# Unique Crashes $\uparrow$} & \textbf{\# Total Crashes $\uparrow$} & \textbf{Coverage $\uparrow$} \\ 
        \specialrule{0.15mm}{0em}{0em}
        \specialrule{0.15mm}{.1em}{0em}
        1 & 1  & 2.0 & 3.0 & \textbf{20968.7} \\
        \hline
        1 & 5  & 3.7 & \textbf{7.0} & 20771.0 \\ 
        \hline
        1 & 10 & \textbf{4.3} & 6.0 & 19879.0 \\
        \specialrule{0.35mm}{0em}{0em}
    \end{tabular}
}
\label{tab:rq2_ucb_reward_settings}
\vspace{-10pt}
\end{table}

\subsubsection{\blue{Effectiveness of LLM-dependent Steps}}
\blue{This section evaluates the accuracy of two LLM-dependent intermediate steps in \MethodName: semantic stage segmentation and LLM-based mutation feasibility analysis.}
\blue{For semantic stage segmentation, we collect all 454 segmentation results produced by the default \MethodName during the experiments in Sec.~\ref{sec:evaluation-rq2_overall_effectiveness} for both the official and newly generated seeds. We randomly sample 20\% (91 results) for manual inspection.}
\blue{We carefully inspect each segmented stage and consider it correct only if both its code content and stage category are correct. For ambiguous cases, we consult the official documentation and reach consensus through discussion. 
The accuracy of each seed is calculated as the proportion of correctly segmented stages.
\textbf{\MethodName achieves an average accuracy of 93.5\% in semantic stage segmentation across the sampled seeds.} The LLM's strong code-understanding capabilities effectively support the segmentation of complex \SimName scripts.} 
\blue{For example, even in a sophisticated robotic grasping scenario with heterogeneous entities and dynamic execution logic, \MethodName correctly segments the code into six semantic stages.
The remaining errors tend to occur in less common code patterns for which the LLM may exhibit limited understanding. For instance, \MethodName misclassifies camera-addition code involving complex initialization and configuration logic as \textit{Interacting} rather than \textit{Adding Sensors}, narrowing the mutation scope (see Table~\ref{tab:stage_object_mapping}).
Future work could inject knowledge relevant to the code context to further enhance accuracy.} \looseness=-1

\begin{table}
\caption{\blue{Evaluation of object-level mutation operators with different object selection strategies.}}
\label{tab:rq3_context_aware_object_selection}
\centering
    \resizebox{\linewidth}{!}
    {
        \begin{tabular}{c|cccccc} 
        \specialrule{0.35mm}{0em}{0em}
        \textbf{Operator}  & \begin{tabular}[c]{@{}c@{}}\textbf{\# Valid \& Cov. Inc.}\\\textbf{Scripts $\uparrow$}\end{tabular} & \begin{tabular}[c]{@{}c@{}}\textbf{\# Valid}\\\textbf{Scripts $\uparrow$}\end{tabular} & \begin{tabular}[c]{@{}c@{}}\textbf{\# Invalid}\\\textbf{Scripts $\downarrow$}\end{tabular} & \begin{tabular}[c]{@{}c@{}}\textbf{\# Unique}\\\textbf{Classes/Pairs$\uparrow$}\end{tabular} & \begin{tabular}[c]{@{}c@{}}\textbf{Coverage}\\\textbf{$\uparrow$}\end{tabular} & \textbf{$p$-value}  \\ 
        \specialrule{0.15mm}{0em}{0em}
        \specialrule{0.15mm}{.1em}{0em}
        \begin{tabular}[c]{@{}c@{}}LLM-based\\Addition\end{tabular}        & {\large\textbf{32.0}}                                                                          & {\large\textbf{39.7}}                  & {\large\textbf{12.8}}                     & {\large 17.0}                                    & {\large 16346.0}                 & 3.19E-3 \\
        \rowcolor{gray!15} 
        \begin{tabular}[c]{@{}c@{}}Context-aware\\Addition\end{tabular}    & {\large 27.7}                                                                                  & {\large 30.0}                          & {\large 20.3}                             & {\large\textbf{27.0}}                            & {\large\textbf{16891.7}}         & -- \\
        \hline
        \begin{tabular}[c]{@{}c@{}}LLM-based\\Replacement\end{tabular}     & {\large 7.7}                                                                                   & {\large 8.0}                           & {\large 41.7}                             & {\large 7.0}                                     & {\large 13920.7}                 & 2.22E-3 \\
        \rowcolor{gray!15} 
        \begin{tabular}[c]{@{}c@{}}Context-aware\\Replacement\end{tabular} & {\large\textbf{24.0}}                                                                          & {\large\textbf{30.3}}                  & {\large\textbf{22.7}}                     & {\large\textbf{23.7}}                            & {\large\textbf{15188.0}}         & -- \\
        \specialrule{0.35mm}{0em}{0em}
        \end{tabular}
    }
\vspace{-10pt}
\end{table}

\blue{LLM-based mutation feasibility analysis assesses whether a requested mutation can produce a valid test case and rejects infeasible mutations to save execution time. A false positive, where an infeasible mutation is deemed feasible, only wastes execution time without otherwise affecting the fuzzing process. In contrast, a false negative may discard a useful or crash-triggering test case. We therefore collect all 139 cases rejected as infeasible by the default \MethodName during the experiments in Sec.~\ref{sec:evaluation-rq2_overall_effectiveness} and randomly sample 20\% (28 cases) for manual inspection. For each case, we carefully validate its mutation feasibility by attempting to construct a test case based on the source code and mutation request.}
\blue{We consider a rejected mutation a false negative if this process yields a valid or crash-triggering test case.}
\blue{\textbf{\MethodName's mutation feasibility judgments are highly consistent with the manual validation results.} Only one false negative is identified among the 28 sampled cases. In most cases, \MethodName can correctly assess mutation feasibility based on the available information.} 
\blue{For example, in a Franka robotic arm control scenario, a mutation requests replacing \textit{Articulation} with \textit{XformPrim}. The original program relies on joint-level capabilities provided by \textit{Articulation} (e.g., degree-of-freedom state management), whereas \textit{XformPrim} only supports general transformation operations (e.g., setting poses). By cross-referencing the source code and API documentation, \MethodName identifies the functional gap and correctly rejects this mutation as infeasible.}
\blue{The sole false negative occurs in a pick-and-place scenario, where a mutation requests removing the \textit{BinFilling} class. Although manually deleting this class along with all its dependent variables yields a valid script, the resulting program retains only an empty \textit{World} shell with no core control logic, providing no useful test outcome. \MethodName conservatively rejects this mutation due to the extensive dependency chain, which exceeds what the LLM can reliably trace in a single reasoning pass. As a future improvement, incorporating static analysis tools could assist this reasoning process.} \looseness=-1

\subsection{RQ3: Ablation Study}
\label{sec:evaluation-rq3_ablation_study}
\looseness=-1
\MethodName incorporates several key components to enhance its effectiveness, i.e., context-aware object selection, multi-level mutation, and UCB-based mutation operator selection. In this section, we evaluate the contribution of each component via ablation studies.
We design four configurations for context-aware object selection, four for mutation operators, and two for operator selection strategy, yielding 10 configurations in total. Each is repeated three times and executed for two hours, resulting in 30 independent runs. 


\subsubsection{Context-aware Object Selection}

We develop LLM-based addition and replacement baselines, in which the LLM directly selects objects from the seed code and the complete list of \SimName classes (detailed prompts are available at~\cite{IcFuzz-GitHub-Repo}).
We compare each baseline with \MethodName retaining only the corresponding operator.
The average results are shown in Table~\ref{tab:rq3_context_aware_object_selection}.
\textbf{Context-aware object selection enables \MethodName to cover more unique classes/pairs and achieve higher code coverage than the LLM-based baseline.}  
In the addition operator, although it generates 4.3 fewer valid and coverage-increasing scripts than the LLM-based method, it covers 10 more unique classes and achieves an additional 545.7 lines of code coverage.
This can be attributed to the LLM's tendency to repeatedly select common classes; for example, \textit{Camera} is selected an average of 5.7 times.
For the replacement operator, context-aware object replacement outperforms the baseline across all metrics. It covers 16.7 more unique object pairs and increases code coverage by 1,267.3 lines.
\blue{The coverage differences are statistically significant for both addition and replacement.}
In contrast, the LLM frequently generates invalid object pairs for replacement. For instance, it replaces \textit{SimulationApp} with \textit{AppFramework} an average of 27.7 times. Since \textit{AppFramework} is a minimal version that lacks essential functionalities~\cite{IsaacSim-NvidiaAPIDoc-SimulationApp}, such replacements fail to produce valid scripts. \looseness=-1



\subsubsection{Mutation Operators}

\begin{table}
\caption{\blue{Ablation study of mutation operators.}}
\centering
    \resizebox{\linewidth}{!}
    {
        \begin{tabular}{c|ccccc} 
        \specialrule{0.35mm}{0em}{0em}
        \textbf{Method}               & \begin{tabular}[c]{@{}c@{}}\textbf{\# Valid \& Cov. Inc.}\\\textbf{Scripts $\uparrow$}\end{tabular} & \begin{tabular}[c]{@{}c@{}}\textbf{\# Valid}\\\textbf{Scripts $\uparrow$}\end{tabular} & \begin{tabular}[c]{@{}c@{}}\textbf{\# Invalid}\\\textbf{Scripts $\downarrow$}\end{tabular} & \begin{tabular}[c]{@{}c@{}}\textbf{Coverage}\\\textbf{$\uparrow$}\end{tabular} & \textbf{$p$-value}  \\ 
        \specialrule{0.15mm}{0em}{0em}
        \specialrule{0.15mm}{.1em}{0em}
        IcFuzz                        & {\large 21.3}                          & {\large 26.7}             & {\large 16.3}               & {\large\textbf{16532.0}}     & -- \\ 
        \hline
        \begin{tabular}[c]{@{}c@{}}IcFuzz w/o\\Object-level Ops\end{tabular}   & {\large\textbf{22.3}}                  & {\large\textbf{27.3}}     & {\large\textbf{12.0}}       & {\large 16043.0}              & 2.57E-2 \\ 
        \hline
        \begin{tabular}[c]{@{}c@{}}IcFuzz w/o\\Operation-level Ops\end{tabular}   & {\large 15.0}                          & {\large 20.0}             & {\large 22.0}               & {\large 15646.0}              & 1.82E-2 \\ 
        \hline
        \begin{tabular}[c]{@{}c@{}}IcFuzz w/o\\Argument-level Ops\end{tabular} & {\large 17.3}                          & {\large 25.7}             & {\large 18.7}               & {\large 15991.3}            & 6.09E-2 \\
        \specialrule{0.35mm}{0em}{0em}
    \end{tabular}
    }
    \label{tab:rq3_op_ablation}
\vspace{-8pt}
\end{table}

We develop three variants of \MethodName, each implemented by individually removing the mutation operators of a specific level. 
The averaged results are shown in Table~\ref{tab:rq3_op_ablation}.

\textbf{Mutation operators at all three levels contribute to improving the coverage achieved by \MethodName.} 
\MethodName achieves the highest code coverage (16,532 lines) compared to the variants, and removing any single level of mutation operators leads to a coverage decrease.
Specifically, \MethodName achieves coverage improvements of 489, 886, and 540.7 lines over the variants without object-, operation-, and argument-level operators, respectively.
\blue{The coverage differences are statistically significant for the variants without object- and operation-level operators, while the variant without argument-level operators yields a $p$-value of 0.0609.}
The benefit stems from the systematic exercise of the hierarchical simulation control enabled by mutations at all three levels.
Regarding the number of coverage-increasing valid scripts, \MethodName generates 6.3 and 4 more such scripts than the variants without operation- and argument-level operators, respectively, indicating that the absence of the corresponding levels of mutation operators negatively affects the generation of effective scripts. 
\MethodName generates slightly fewer valid scripts with coverage increase (by 1) than the variant without object-level operators, since object-level mutations are inherently difficult and result in more invalid scripts (by 4.3) on average. However, enabling object-level operators facilitates exploration at an additional granularity, ultimately allowing IcFuzz to achieve the highest overall coverage. \looseness=-1


\subsubsection{Mutation Operator Selection}
\begin{table}
\caption{\blue{Evaluation of mutation operator selection strategies.}}
\label{tab:rq3_ucb_vs_random}
\centering
    \resizebox{\linewidth}{!}
    {
        \begin{tabular}{c|ccccc} 
        \specialrule{0.35mm}{0em}{0em}
        \textbf{Method}          & \begin{tabular}[c]{@{}c@{}}\textbf{\# Valid \& Cov. Inc.}\\\textbf{Scripts $\uparrow$}\end{tabular} & \begin{tabular}[c]{@{}c@{}}\textbf{\# Valid}\\\textbf{Scripts $\uparrow$}\end{tabular} & \begin{tabular}[c]{@{}c@{}}\textbf{\# Invalid}\\\textbf{Scripts $\downarrow$}\end{tabular} & \begin{tabular}[c]{@{}c@{}}\textbf{Coverage}\\\textbf{$\uparrow$}\end{tabular} & \textbf{$p$-value}         \\ 
        \specialrule{0.15mm}{0em}{0em}
        \specialrule{0.15mm}{.1em}{0em}
        IcFuzz (UCB)    & {\large\textbf{21.3}}               & {\large\textbf{26.7}}    & {\large\textbf{16.3}}      & {\large\textbf{16532.0}}  & -- \\ 
        \hline
        IcFuzz (Random) & {\large 16.0}                       & {\large 22.3}            & {\large 19.7}              & {\large 15331.3}          & 3.55E-4 \\
        \specialrule{0.35mm}{0em}{0em}
        \end{tabular}
    }
\vspace{-6pt}
\end{table}
We develop a variant of \MethodName that selects mutation operators uniformly at random as a baseline. The average results are reported in Table~\ref{tab:rq3_ucb_vs_random}.
\textbf{UCB-based mutation operator selection allows \MethodName to generate more valid scripts and achieve higher code coverage than random selection.} 
Specifically, \MethodName with UCB generates 4.4 more valid scripts and 5.3 more valid coverage-increasing scripts, while producing 3.4 fewer invalid scripts than random selection. The code coverage increases by 1,200.7 lines, \blue{and the difference is statistically significant.} \looseness=-1 

%% file: section/05-threats.tex
\looseness=-1
Threats to internal validity mainly arise from implementation errors and the randomness of fuzzing. 
To mitigate potential errors, we implemented the baselines using the replication package provided by GzFuzz and the official repository of Atheris. Only necessary modifications were made to GzFuzz for adapting to Isaac Sim. Additionally, we carefully reviewed the code of \MethodName and inspected intermediate outputs to ensure correctness.
Regarding randomness, we mitigate its impact by repeating each experiment in RQ2 and RQ3 under identical settings and reporting the average results.

Threats to external validity mainly concern the generalizability of the results. 
\blue{Not all simulation cases strictly conform to the summarized semantic stages; instead, these stages are intended to capture critical and representative simulation workflows.} 
\blue{Although \MethodName is currently evaluated on \SimName, its methodology is designed around characteristics common to existing robotics simulators to improve generalizability. The potentially transferable aspects of \MethodName lie in the semantic stage-guided design, i.e., lifecycle-style stage segmentation, stage constraints for semantic validity, multi-level mutation, and feedback-based mutation scheduling, which target common properties of script-driven robotics simulators. For instance, scripts in Genesis~\cite{Genesis}, MuJoCo~\cite{MuJoCo}, and Webots~\cite{Webots} can likewise be divided into stages such as initialization and object addition, and share common object types representing simulation elements (e.g., sensors and materials). Porting \MethodName to another simulator requires engineering effort to adapt several components, including seed pool construction, stage-to-object mapping, API documentation crawling, simulator-specific prompt rules, and the execution harness. We leave the adaptation and evaluation of \MethodName on other robotics simulators to future work.}






%% file: section/06-related_work.tex
\textbf{Bug studies in robotic software.} 
Prior work has studied robotic software bugs from various perspectives.
Empirical studies have characterized bugs in the Robot Operating System (ROS)~\cite{ROS2}, including a curated dataset~\cite{DBLP:journals/ese/TimperleyHSDW24-ROBUST221bugs}, dependency bugs~\cite{DBLP:conf/icse/Fischer-Nielsen20-ROSDependency}, and interaction bugs~\cite{DBLP:journals/corr/abs-2507-10235-iBug}.
For testing, RoboFuzz~\cite{DBLP:conf/sigsoft/KimK22-RoboFuzz} leverages domain oracles and feedback to identify semantic correctness bugs in ROS through fuzzing, while R2D2~\cite{DBLP:conf/issta/ShenLXSWGS024-R2D2FuzzingCallback} guides fuzzing in ROS using callback traces.
PhyFu~\cite{DBLP:conf/kbse/XiaoLW23-Phyfu} mutates physical states to detect physical law violations in physics simulation engines.
GzFuzz~\cite{DBLP:journals/pacmse/RenLLQXJ25-GzFuzz} is the SOTA approach for fuzzing the robotics simulator Gazebo. It does not cover finer-grained simulation control and enables only limited exploration when applied to Isaac Sim, given its more complex architecture. 
In contrast, \MethodName targets standalone scripts that enable precise simulation control in \SimName and achieves more systematic testing through the proposed techniques (e.g., multi-level mutation).


\blue{\textbf{Conventional automated test generation.}
Prior work has explored structured test input generation through grammar-based and search-based methods.
Grammar-based fuzzing (e.g., Superion~\cite{DBLP:conf/icse/WangCWL19-Superion} and NAUTILUS~\cite{DBLP:conf/ndss/AschermannFHJST19-NAUTILUS}) generates or mutates inputs according to predefined rules (e.g., grammars and templates).
These extensively hand-crafted rules mainly capture syntactic structures and are difficult to extend to simulator-specific semantics (e.g., inter-dependencies among simulation elements). 
Representative search-based tools synthesize tests by optimizing objectives such as code coverage for individual classes, functions, or modules (e.g., EvoSuite~\cite{DBLP:conf/sigsoft/FraserA11-EvoSuite} and Pynguin~\cite{DBLP:conf/icse/LukasczykF22-Pynguin}).
In contrast, \MethodName fuzzes the entire simulator by generating syntactically and semantically valid test scripts.}

\looseness=-1
\textbf{LLMs for Software Testing.}
Recent studies have applied LLMs to diverse software testing targets with tailored designs.
TitanFuzz~\cite{DBLP:conf/issta/DengXPY023-titanfuzz} generates and mutates test programs for deep learning library fuzzing, GPTDroid~\cite{DBLP:conf/icse/0025C0CWCW024-GPTDroid} formulates mobile GUI testing as a multi-turn Q\&A task, and Fuzz4All~\cite{DBLP:conf/icse/XiaPTP024-fuzz4all} achieves universal fuzzing across languages via iterative LLM-driven generation. 
While effective, these approaches assume that testing targets are relatively common and that LLMs have acquired sufficient knowledge during pre-training~\cite{DBLP:conf/issta/DengXPY023-titanfuzz,DBLP:conf/icse/0025C0CWCW024-GPTDroid,DBLP:conf/icse/XiaPTP024-fuzz4all}. 
In contrast, \MethodName targets Isaac Sim, which involves highly specialized robotics domain knowledge and a rapidly evolving software ecosystem~\cite{IsaacSim-GitHubPage}, largely absent from pre-training.
Thus, directly generating or mutating standalone scripts with LLMs is often ineffective.
Instead of the costly fine-tuning to inject domain knowledge as adopted in
prior work~\cite{DBLP:conf/vts/TarekSSF25-finetune,DBLP:conf/kbse/SunYWWJZ23-finetune,DBLP:conf/icse/LiuCWCHHW23-finetune}, \MethodName enhances test effectiveness by strictly guiding, constraining, and validating LLM outputs without updating the model.




%% file: section/07-conclusion.tex
We propose \MethodName, the first fuzzing approach dedicated to Isaac Sim. 
\MethodName employs LLMs to segment seed scripts into semantic stages and uses the resulting stage information to guide context-aware object selection. 
It utilizes multi-level mutation operators to systematically exercise the hierarchical simulation control. \MethodName further schedules mutation operators adaptively based on execution feedback to efficiently explore the simulation state space. 
Experimental results demonstrate that \MethodName outperforms existing SOTA fuzzing approaches in both code coverage and \blue{bug detection}. 
Our approach also provides insights into testing other robotics simulators. 

